\documentclass[10pt,journal,compsoc]{IEEEtran}
\ifCLASSOPTIONcompsoc
  \usepackage[nocompress]{cite}
\else
  \usepackage{cite}
\fi

\ifCLASSINFOpdf
  \usepackage[pdftex]{graphicx}
  \graphicspath{{../pdf/}{../jpeg/}}
  \DeclareGraphicsExtensions{.pdf,.jpeg,.png}
\else
  \usepackage[dvips]{graphicx}
  \graphicspath{{../eps/}}
  \DeclareGraphicsExtensions{.eps}
\fi

\ifCLASSOPTIONcompsoc
 \usepackage[caption=false,font=footnotesize,labelfont=sf,textfont=sf]{subfig}
\else
 \usepackage[caption=false,font=footnotesize]{subfig}
\fi

\usepackage{url}
\usepackage[breaklinks,colorlinks]{hyperref}
\hypersetup{colorlinks=true}
\usepackage{ragged2e}
\usepackage{xcolor}
\usepackage{color}
\usepackage{colortbl}
\definecolor{figure_head}{gray}{.85}
\usepackage{threeparttable}
\usepackage{multirow}
\usepackage{diagbox}
\usepackage{tikz}
\usepackage{amsmath}
\usepackage{amssymb}
\usepackage{makecell}
\usepackage{pifont}

\begin{document}
\title{
Deep Image Matting: A Comprehensive Survey
}

\author{Jizhizi Li,
        Jing Zhang,~\IEEEmembership{Member,~IEEE},
        Dacheng Tao,~\IEEEmembership{Fellow,~IEEE}
    \thanks{Jizhizi Li, Jing Zhang, and Dacheng Tao are with the School of Computer Science, Faculty of Engineering, The University of Sydney, Australia (e-mail: jili8515@uni.sydney.edu.au; jing.zhang1@sydney.edu.au; dacheng.tao@gmail.com)}
}

\markboth{Journal of \LaTeX\ Class Files,~Vol.~x, No.~x, x~x}%
{Shell \MakeLowercase{\textit{et al.}}: Bare Demo of IEEEtran.cls for Computer Society Journals}

\IEEEtitleabstractindextext{
\begin{abstract}
\justifying
Image matting refers to extracting precise alpha matte from natural images, and it plays a critical role in various downstream applications, such as image editing. Despite being an ill-posed problem, traditional methods have been trying to solve it for decades. The emergence of deep learning has revolutionized the field of image matting and given birth to multiple new techniques, including automatic, interactive, and referring image matting. This paper presents a comprehensive review of recent advancements in image matting in the era of deep learning. We focus on two fundamental sub-tasks: auxiliary input-based image matting, which involves user-defined input to predict the alpha matte, and automatic image matting, which generates results without any manual intervention. We systematically review the existing methods for these two tasks according to their task settings and network structures and provide a summary of their advantages and disadvantages. Furthermore, we introduce the commonly used image matting datasets and evaluate the performance of representative matting methods both quantitatively and qualitatively. Finally, we discuss relevant applications of image matting and highlight existing challenges and potential opportunities for future research. We also maintain a public repository to track the rapid development of deep image matting at {\url{https://github.com/JizhiziLi/matting-survey}}.
\end{abstract}
}

\maketitle
\IEEEdisplaynontitleabstractindextext
\IEEEpeerreviewmaketitle

\IEEEraisesectionheading{\section{Introduction}\label{sec:introduction}}

\IEEEPARstart{I}{mage} matting is a fundamental computer vision problem that involves extracting the precise alpha matte of foreground objects from arbitrary natural images. This technique is particularly suitable for categories that have very fine details, such as humans~\cite{dapm,shm,p3mj}, animals~\cite{gfm}, and plants~\cite{aim}. Image matting serves as an essential procedure for various downstream tasks, including promotional advertising in practical e-commerce platforms, image editing for daily life entertainment, background replacing for online video conferences, and metaverse applications such as virtual reality and game industries.

Image matting, initially referred to as the \textit{travelling-matte problem} in the film-making industry~\cite{Beyer1965TravelingMattePA, Porter1984CompositingDI}, has been a subject of research for several decades. However, due to the ill-posed nature of the problem, traditional methods have relied on various auxiliary user inputs to alleviate the challenge. These inputs include trimap~\cite{wang2007optimized, Sun2004PoissonM, Li2018PatchAM, aksoy2017designing} and scribble~\cite{Levin2006ACS, zheng2008fuzzymatte, wang2005iterative, Bai2007AGF}, or specific conditions such as blue-screen~\cite{Beyer1965TravelingMattePA, Smith1996BlueSM}. Previous methods for image matting can be divided into two main streams. The first stream~\cite{Chuang2001ABA} employs color sampling to estimate the unknown pixels from neighbouring ones, while the latter stream defines an affinity matrix~\cite{zheng2008fuzzymatte} to propagate the known foreground and background to the transition area. Both of the two streams rely on low-level color or structure features, which limit their ability to distinguish foreground details from complex backgrounds, resulting in sensitivity to the size of the unknown area and fuzzy boundaries~\cite{ruzon2000alpha, shahrian2013improving}.

In recent years, deep learning has emerged as a new approach to solving various computer vision problems including image matting~\cite{zhang2020empowering}. Researchers have designed new solutions based on deep convolutional neural networks (CNNs) that are capable of learning discriminative features~\cite{Cho2016NaturalIM,dapm}. These approaches typically follow the conventional settings of using auxiliary user input to add a constraint to the ill-posed problem, which can include trimap~\cite{dim,gca,lu2019indices,Forte2020FBA,Liu2021TripartiteIM}, scribble~\cite{Yang2020SmartSF}, background image~\cite{backgroundmatting}, coarse map~\cite{Cho2016NaturalIM,mgmatting}, or even text description~\cite{rim}. Some methods concatenate the image and auxiliary inputs and pass them through a single-stage network~\cite{dim,Wang2018DeepPB,lu2019indices,hou2019context} with or without an auxiliary module, while others leverage a multi-stream structure for task-aware optimization~\cite{Tang2019LearningBasedSF,cai2019disentangled}. Although these approaches have produced results with better details than traditional solutions, they still require a significant amount of manual effort to provide auxiliary inputs. To address this limitation, more flexible auxiliary input options have been explored, such as user-click~\cite{NEURIPS2018_653ac11c,Wei2020ImprovedIM,Ding2022DeepII} or text description~\cite{rim}. Furthermore, the success of transformer~\cite{NIPS2017_3f5ee243} and Vision Transformer~\cite{vit,Liu2021SwinTH,vitae} has also led to the introduction of these structures in image matting~\cite{Park2022MatteFormerTI,rim,dai2022boosting}.

\begin{figure*}[t]
    \centering
    \includegraphics[width=\linewidth]{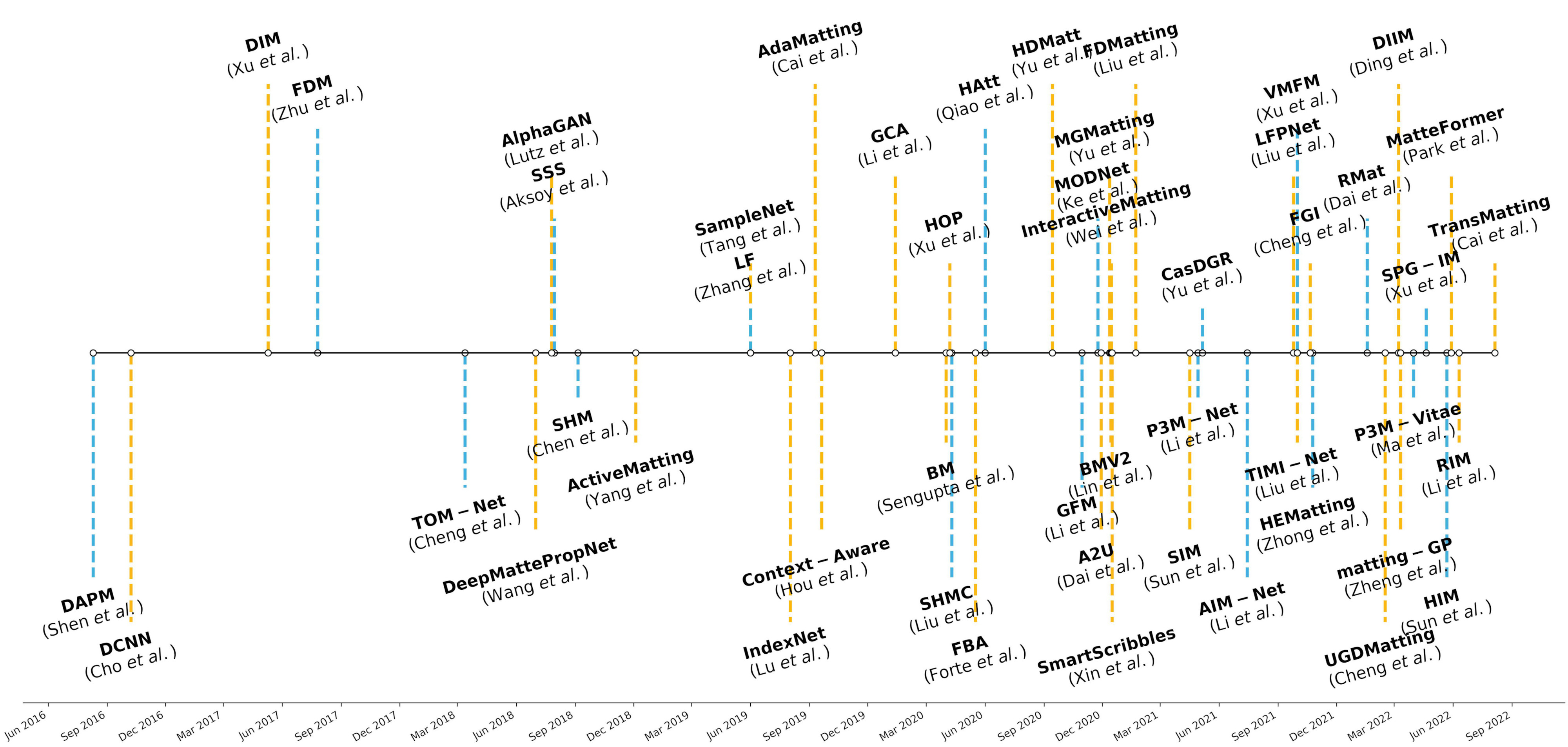}
    \caption{The timeline presents a historical overview of deep learning-based image matting methods. The blue lines indicate the auxiliary input-based matting methods, while the yellow lines indicate the automatic matting methods.
    }
    \label{fig:timeline}
\end{figure*}

In contrast to the above methods, automatic image matting methods aim to predict the foreground of an image without requiring any user intervention~\cite{dapm,Zhu2017FastDM}. These methods typically predict specific salient foreground objects, which are implicitly defined by the training dataset, \eg, human~\cite{dapm,shm,shmc,modnet,p3m}, animal~\cite{gfm}, or composite foregrounds~\cite{hatt,lf,Qiao2020MultiscaleIA}. Recently, AIM~\cite{aim} categorizes the foreground objects in nature images into three kinds, \ie, salient opaque, salient transparent, and non-salient, and processes them through a single model. As per the network structure, automatic matting methods can be divided into three groups: one-stage network with global guidance~\cite{Yu2021CascadeIM}, sequential segmentation and matting network~\cite{dapm,shm}, and parallel multi-task network~\cite{gfm,aim} to model subtasks explicitly. Recently, some methods have also introduced transformer structures~\cite{zhang2023vitaev2,zhang2023vision} into automatic image matting~\cite{p3mj}. These methods use the self-attention mechanism to capture long-range dependencies and context information, which can improve the accuracy of the predictions.

To facilitate the training and evaluation of deep learning-based methods, many datasets have been proposed over the years. The first such dataset was proposed by alphamatting~\cite{rhemann2009perceptually}, containing 27 images and a benchmark website. Since then, numerous datasets have been proposed by researchers, containing different content categories, image resolutions, composition styles, and annotation strategies~\cite{dim,dapm,shm,lf,hatt,gfm,aim,p3m,modnet,mgmatting}. Due to the laborious and costly labelling process, many early works~\cite{dim,lf,hatt} either use chroma keying to extract foregrounds and alpha blending~\cite{Prez2003PoissonIE} to compose synthetic images with backgrounds from MS COCO~\cite{coco} and Pascal VOC~\cite{everingham2010pascal} or adopt existing matting methods close-form~\cite{Levin2006ACS}, and KNN~\cite{chen2013knn} to compute matte labels. Although these early datasets provide valuable training data and facilitate the development of deep learning-based methods, the composition artifacts and limited resolution could mislead trained models and result in poor generalization abilities on natural images~\cite{gfm,modnet}. Recently, some researchers have proposed high-resolution natural image datasets along with manually labeled alpha mattes to resolve these issues~\cite{gfm,aim,modnet,mgmatting}. Furthermore, privacy-preserving dataset~\cite{p3m} has also been established to address ethical issues in portrait image matting.

In this article, we present a comprehensive review of recent advancements in image matting achieved through the use of deep learning techniques. We discuss two broad categories of methods: auxiliary input-based matting and automatic matting. We also cover the different datasets used for training and testing these methods, evaluate their performance, and discuss the potential applications and future work. We have reviewed 69 papers from top-tier conferences and journals and compiled a timeline of the developments in deep learning-based image matting in Figure~\ref{fig:timeline}, highlighting the growing intensity of research in this area. To keep pace with the fast development, we have established a public repository\footnote{\url{https://github.com/JizhiziLi/matting-survey}} that provides an up-to-date record of recent advancements. In summary, this review aims to provide researchers and practitioners in this field with a comprehensive understanding of the current state-of-the-art techniques, their performance, related applications, and potential avenues for future work.

\subsection{Contributions of this Survey}

Compared with prior matting surveys that predominantly focus on traditional methods~\cite{piccardi2004background,Wang2007ImageAV,Boda2018ASO,Li2019ASO,Yao2017ACS}, this paper provides a comprehensive review of image matting methods, with a specific emphasis on those utilizing deep learning techniques, making it unique and important in both scope and depth. The contributions of this survey can be summarized as follows:

$\bullet$ We make the first attempt to provide a comprehensive summary and literature review of deep learning-based image matting methods.

$\bullet$ We examine the progress made in deep image matting from several perspectives, including input and output forms, network structures, and benchmark datasets.

$\bullet$ We offer a fair and comprehensive evaluation of the performance of deep image matting methods, while also analyzing their model complexity.

$\bullet$ We discuss promising applications of image matting and highlight the challenges and potential opportunities for further research.

\subsection{Relationship to Other Surveys}

In order to provide a comprehensive understanding of the research landscape and contextualize our analysis of deep image matting methods, we suggest consulting several excellent previous surveys that focus on image matting~\cite{Boda2018ASO,Li2019ASO,Yao2017ASO,Yao2017ACS} and related research topics, \eg, video matting~\cite{Wang2007ImageAV,dhivyaavideo}, image composition~\cite{Niu2021MakingIR}, semantic segmentation~\cite{Minaee2020ImageSU,GarciaGarcia2018ASO,Lateef2019SurveyOS,Zhao2017ASO,Thoma2016ASO}, and salient object detection~\cite{Borji2014SalientOD,Han2018AdvancedDT,Wang2019SalientOD}. Previous surveys on image matting have primarily focused on traditional solutions, examining aspects such as methodologies~\cite{Li2019ASO,Wang2007ImageAV}, and pre-processing strategies~\cite{Yao2017ASO}. These methods have been categorized as sampling-based~\cite{Chuang2001ABA}, affinity-based~\cite{Levin2006ACS,Sun2004PoissonM,grady2005random}, or a combination of the two~\cite{wang2005iterative,wang2007optimized,guan2006easy}. Other works have explored downstream applications of image matting, such as video matting~\cite{chuang2002video,Bai2007AGF,mcguire2005defocus}, environment matting~\cite{wexler2002image,zongker1999environment,duan2011fast}, shadow matting~\cite{wu2007natural,chuang2003shadow}, composition~\cite{Wang2006SimultaneousMA}, and harmonization~\cite{Sunkavalli2010MultiscaleIH}. While some surveys on relevant research topics have provided comprehensive analyses of methodologies and datasets~\cite{dhivyaavideo,Minaee2020ImageSU,Borji2014SalientOD}, they have not investigated image matting in depth. In contrast, this paper surveys large amounts of relevant image matting methods, with a particular focus on those based on deep learning.

\subsection{Organization}

The remainder of this paper is structured as follows. In Section~\ref{sec:preliminary}, we present some preliminary background information regarding the problem statement, taxonomy, and traditional solutions. Section~\ref{sec:auxiliary} reviews auxiliary input-based matting models from the perspective of various input forms. Section~\ref{sec:automatic} discusses automatic matting methods that use either sequential or parallel structures. Section~\ref{sec:dataset} provides an overview of both composite and natural matting datasets. In Section~\ref{sec:performance}, we conduct a performance evaluation and analysis of representative matting methods on several benchmark datasets. Subsequently, in Section~\ref{sec:applications}, we present the relevant applications of deep image matting, followed by a discussion of the challenges and opportunities in Section~\ref{sec:challenges}. Finally, we conclude the review in Section~\ref{sec:conclusion}.

\section{Preliminary}
\label{sec:preliminary}

\subsection{Problem Statement}

Image matting, which refers to the precise extraction of the soft matte from foreground objects in arbitrary images, has been extensively studied for several decades. The concept was initially introduced by Beyer~\cite{Beyer1965TravelingMattePA} in the 1960s and later mathematically formalized by Porter and Duff~\cite{Porter1984CompositingDI} in the 1980s. Subsequently, researchers have referred to this problem as \textit{pulling matte}, \textit{matte creation} \cite{smith1995alpha}, or \textit{chromakey} \cite{Smith1996BlueSM}. The problem has been studied in conjunction with the problem of \textit{digital composition}~\cite{fielding1974technique,blinn1994compositing,smith1995image}.

\begin{equation}
\label{eqa:matting}
I_i = \alpha_iF_i+(1-\alpha_i)B_i, \quad\quad\quad \alpha_i \in \left[ 0,1 \right].
\end{equation}

The image matting process can be described mathematically using Eq.~\eqref{eqa:matting}, where $I$ represents the input image, $F$ represents the foreground image, and $B$ represents the background image. In this equation, the color of the $i$th pixel is approximated as a convex combination of the corresponding foreground and background colors. The opacity of the pixel in the foreground is denoted by $\alpha_i$, which ranges from 0 to 1. If $\alpha_i$ equals 1, the pixel is classified as a \textit{pure foreground} or \textit{definite foreground}. Conversely, if $\alpha_i$ equals 0, it is classified as a \textit{pure background} or \textit{definite background}. Pixels with opacity values ranging from 0 to 1 are known as the \textit{unknown region} or \textit{transition area}. Since $I$ is a three-channel RGB image, image matting is defined by three equations and seven unknown variables, resulting in a heavily ill-posed problem that requires in-depth research study. 

\subsection{Taxonomy}

In this section, we discuss deep learning-based matting methods using a taxonomy that encompasses the input modality, matting target, and methodology. Table~\ref{tab:papers} summarizes the reviewed papers, ordered by their release dates.

\subsubsection{Input Modality}

Matting methods adopt different input modalities that suit both automatic and auxiliary-based methodologies. For methods that require auxiliary information to constrain the solution space, the input modalities can be further categorized into various types, including RGB image with trimap~\cite{Wang2018DeepPB,bmvcLutzAS18,lu2019indices,hou2019context}, RGB image with background images~\cite{backgroundmatting,backgroundmattingv2}, RGB image with coarse maps~\cite{mgmatting,Cho2016NaturalIM}, RGB image with user click~\cite{Wei2020ImprovedIM,Ding2022DeepII}, RGB image with flexible inputs~\cite{Cheng2021DeepIM}, and RGB image with text descriptions~\cite{rim}. Among these input modalities, the \textit{trimap} is the most commonly used, which is a three-class map that indicates the pure foreground, pure background, and unknown region~\cite{Wang2007ImageAV}. Each type of auxiliary input provides varying degrees of aid in easing the burden of the ill-posed matting problem. On the other hand, for automatic methods~\cite{shm,Zhu2017FastDM,Chen2018TOMNetLT,gfm} designed for automatic industrial applications, the input modality is a single RGB image that can be either a composite one or a natural one.

\subsubsection{Matting Target}

In general, there are no restrictions on the types of foreground objects that can be processed as matting targets. However, the (generalization) ability of deep learning-based image matting methods is limited by the foreground types available in the training dataset. Several studies have focused on human (portrait) matting, as portraits are a prevalent subject in the matting task~\cite{dapm, p3mj, shmc, backgroundmatting, modnet}. Other researchers have explored foreground types that have meticulous details, such as animals~\cite{gfm}, or transparent objects~\cite{Chen2018TOMNetLT,cai2022transmatting}. In~\cite{aim}, the matting targets are categorized into three groups: salient opaque foregrounds, salient transparent foregrounds, and non-salient foregrounds, which represent a typical taxonomy of matting targets.

\subsubsection{Methodology}

The taxonomy of matting methodologies can be approached from two perspectives, namely auxiliary input-based methods and automatic methods. With regard to the former, the methodologies can be classified into three categories. Firstly, a single one-stage CNN is used to directly map the concatenation of the input image and the auxiliary input to the alpha matte~\cite{dim, Forte2020FBA, gca}. Secondly, a one-stage CNN is used with modules carefully designed to make use of the rich features from the auxiliary input through the side branch~\cite{Wang2018DeepPB, sim, Ding2022DeepII}. Thirdly, parallel two- or multi-stream structures are utilized to decompose the matting task into explicit sub-tasks~\cite{cai2019disentangled, backgroundmatting, Yu_Xu_Huang_Zhou_Shi_2021}. Some of these methodologies require an additional refiner to further refine the predicted alpha matte~\cite{dim, backgroundmattingv2}. More details about these methodologies will be discussed in Section~\ref{sec:auxiliary}.

For automatic matting methods, there are also three main methodologies. The first is a one-stage structure, which can optionally include a global module~\cite{hatt,Yu2021CascadeIM} as guidance to predict the matte directly from a single input. The second is a sequential two-step structure, where an intermediate segmentation mask or trimap is generated first, and then combined with the initial input to produce the final alpha matte~\cite{dapm, Zhu2017FastDM, shm, shmc}. The third methodology is a parallel two- or multi-stream structure, which decomposes the matting task into several sub-tasks such as foreground and background or global semantic mask~\cite{lf} and local details~\cite{gfm, p3m, aim}. Some of these methodologies also require a refiner to refine the alpha matte~\cite{Chen2018TOMNetLT}. More details about these methodologies will be discussed in Section~\ref{sec:automatic}.

\begin{table*}
\centering
\caption{Summary of image matting methods organized according to the year of publication, the publication venue, input modality, automaticity, matting target, architecture, train set, and test set. Please note that the list of papers is chronologically ordered.
}
\vspace{-5pt}
\label{tab:papers}
\begin{threeparttable}
\resizebox{1\textwidth}{!}{
\setlength\tabcolsep{6pt}
\renewcommand\arraystretch{1.15}
\begin{tabular}{|c|r|c||c|c|c|c|c|c|}
\hline
\rowcolor{figure_head}
Year &Method &Pub. & Modality & Automatic & Target & Architecture  & Train Set  & Test Set \\
\hline
\hline
\multirow{2}{*}{\rotatebox{90}{2016}} & DAPM~\cite{dapm} & ECCV & RGB &\checkmark &  human& Sequential two-step CNN & DAPM-2k~\cite{dapm}& DAPM-2k~\cite{dapm}\\
 & DCNN~\cite{Cho2016NaturalIM} & ECCV & RGB-Coarse &  & object & One-stage CNN  & AlphaMatting~\cite{rhemann2009perceptually} &AlphaMatting~\cite{rhemann2009perceptually} \\
\hline
\multirow{2}{*}{\rotatebox{90}{2017}} & DIM~\cite{dim} & CVPR & RGB-Trimap & & object & One-stage CNN+Refine & DIM-481~\cite{dim}& DIM-481~\cite{dim}\\
 & FDM~\cite{Zhu2017FastDM} & MM & RGB & \checkmark & human & Sequential two-step CNN  & DAPM-2k~\cite{dapm} &DAPM-2k~\cite{dapm} \\
 \hline
 \multirow{6}{*}{\rotatebox{90}{2018}} & TOM-Net~\cite{Chen2018TOMNetLT} & CVPR & RGB & \checkmark & transparent & Sequential two-step CNN + Refine & TOM-876~\cite{Chen2018TOMNetLT}& TOM-876~\cite{Chen2018TOMNetLT}\\
 & DMPN~\cite{Wang2018DeepPB} & IJCAI & RGB-Trimap&&object & One-stage CNN & DMPN-46~\cite{Wang2018DeepPB}&DMPN-46~\cite{Wang2018DeepPB}\\
 & AlphaGAN~\cite{bmvcLutzAS18} & BMVC & RGB-Trimap&&object & One-stage GAN & DIM-481~\cite{dim}&DIM-481~\cite{dim}\\
& SSS~\cite{sss} & TOG & RGB&\checkmark&object & Sequential two-stage structure & COCO-Stuff~\cite{caesar2018coco}&DIM-481~\cite{dim}\\
& SHM~\cite{shm} & MM & RGB&\checkmark&human & Sequential two-step CNN & SHM-35k~\cite{shm}&SHM-35k~\cite{shm}\\
& ActiveMatting~\cite{NEURIPS2018_653ac11c} & NeurIPS & RGB-Click&&object & One-stage RNN & DAPM-2k~\cite{dapm}&DAPM-2k~\cite{dapm}\\
 \hline
 \multirow{5}{*}{\rotatebox{90}{2019}} &LF~\cite{lf}& CVPR & RGB & \checkmark & object & Sequential two-stage CNN  &DIM-481~\cite{dim}, LF-257~\cite{lf}&DIM-481~\cite{dim}, LF-257~\cite{lf} \\
  &SampleNet~\cite{Tang2019LearningBasedSF}& CVPR & RGB-Trimap & & object & Parallel three-stream CNN &DIM-481~\cite{dim} & DIM-481~\cite{dim}\\
 & IndexNet~\cite{lu2019indices} & ICCV & RGB-Trimap &  & object & One-stage CNN & DIM-481~\cite{dim}&DIM-481~\cite{dim} \\
 & AdaMatting~\cite{cai2019disentangled} & ICCV & RGB-Trimap &  &object & Parallel two-stream CNN + Refine & DIM-481~\cite{dim} & DIM-481~\cite{dim}\\
 & Context-Aware~\cite{hou2019context}& ICCV & RGB-Trimap & & object & Two-stream CNN & DIM-481~\cite{dim} & DIM-481~\cite{dim}\\
 \hline
 \multirow{14}{*}{\rotatebox{90}{2020}} & GCA~\cite{gca} & AAAI & RGB-Trimap& & object & One-stage CNN & DIM-481~\cite{dim} & DIM-481~\cite{dim}\\
 & BM~\cite{backgroundmatting} & CVPR & RGB-Background&& human & Parallel four-stream CNN & DIM-481~\cite{dim}, BM-Video~\cite{backgroundmatting} & DIM-481~\cite{dim}, BM-Video~\cite{backgroundmatting}\\
 & HOP~\cite{Li2020HierarchicalOP} & arXiv & RGB-Trimap & & object & Parallel two-stream CNN &DIM-481~\cite{dim} & DIM-481~\cite{dim} \\
 & SHMC~\cite{shmc} & CVPR & RGB & \checkmark & human & Sequential two-stage CNN & SHMC-10k~\cite{shmc} & SHMC-10k~\cite{shmc}\\
 & FBA~\cite{Forte2020FBA} & arXiv & RGB-Trimap & &object & One-stage CNN & DIM-481~\cite{dim} & DIM-481~\cite{dim}\\
 & HAtt~\cite{hatt} & CVPR & RGB & \checkmark & object & One-stage CNN & DIM-481~\cite{dim}, HATT-646~\cite{hatt} & DIM-481~\cite{dim}, HATT-646~\cite{hatt}\\
 & HDMatt~\cite{Yu_Xu_Huang_Zhou_Shi_2021} & AAAI &RGB-Trimap &  & object & Parallel two-stream CNN &DIM-481~\cite{dim}&DIM-481~\cite{dim}, DAPM-2k~\cite{dapm} \\
 & GFM~\cite{gfm} & IJCV & RGB &\checkmark&human, animal & Parallel two-stream CNN & AM-2k~\cite{gfm}, PM-10k~\cite{gfm} & AM-2k~\cite{gfm}, PM-10k~\cite{gfm}\\
 & MODNet~\cite{modnet} & AAAI &RGB &\checkmark& human & Parallel two-stream CNN & SPD~\cite{sps}, PPM-3000~\cite{modnet}& PPM-100~\cite{modnet}\\
 & A2U~\cite{dai2021learning} & CVPR &RGB-Trimap & & object & One-stage CNN &DIM-481~\cite{dim},HATT-646~\cite{hatt}& DIM-481~\cite{dim},HATT-646~\cite{hatt}\\
 & MGMatting~\cite{mgmatting} & CVPR & RGB-Coarse& & human & One-stage CNN &DIM-481~\cite{dim} & \makecell[c]{DIM-481~\cite{dim}, RWP636~\cite{mgmatting}\\HATT-646~\cite{hatt}}\\
 & InteractiveMatting~\cite{Wei2020ImprovedIM} & CVPR & RGB-Click&  & object & Parallel two-stream CNN &DIM-481~\cite{dim}&DIM-481~\cite{dim} \\ 
 & SmartScribbles~\cite{Yang2020SmartSF} & TOMM & RGB-Scribble&  & object & One-stage CNN &DAPM~\cite{dapm}, DIM-481~\cite{dim}&DAPM~\cite{dapm},DIM-481~\cite{dim} \\ 
 & BMV2~\cite{backgroundmattingv2} & CVPR & RGB-Background& & human& One-stage CNN + Refiner & \makecell[c]{DIM-481~\cite{dim}, HATT-646~\cite{hatt},\\VM-24~\cite{backgroundmattingv2}, PhotoMatte13k~\cite{backgroundmattingv2}}& PhotoMatte85~\cite{backgroundmattingv2}\\
 \hline
 \multirow{13}{*}{\rotatebox{90}{2021}}& FDMatting~\cite{Liu2021TowardsEF} &WACV& RGB-Trimap&& object & Two-stream CNN  & DIM-481~\cite{dim} & DIM-481~\cite{dim}\\
 & SIM~\cite{sim} &CVPR& RGB-Trimap& & object &  One-stage CNN & SIM~\cite{sim} &SIM~\cite{sim}, DIM-481~\cite{dim}\\
& P3M-Net~\cite{p3m} &MM&RGB &\checkmark& human & Parallel two-stream CNN & P3M-10k~\cite{p3m} & P3M-10k~\cite{p3m}\\
& CasDGR~\cite{Yu2021CascadeIM} &ICCV&RGB &\checkmark& object &Multi-scale multi-stage CNN & DIM-481~\cite{dim}& DIM-481~\cite{dim}\\
 & AIM-Net~\cite{aim} &IJCAI&RGB &\checkmark& object & Parallel two-stream CNN & \makecell[c]{ DUTS~\cite{wang2017learning}, AM-2k~\cite{gfm}, \\ DIM-481~\cite{dim}, HATT-646~\cite{hatt}}
 & AIM-500~\cite{aim} \\
 & LFPNet~\cite{Liu2021LongRangeFP} &MM&RGB-Trimap &&object & Parallel two-stream CNN &DIM-481~\cite{dim} & DIM-481~\cite{dim}\\
 & VMFM~\cite{Xu2021VirtualMS} &ICCV&RGB &\checkmark&human-object & Sequential two-stage CNN & \makecell[c]{DIM-481~\cite{dim}, HATT-646~\cite{hatt}\\LFM40k~\cite{Xu2021VirtualMS}, UFM75k~\cite{Xu2021VirtualMS}} & DIM-481~\cite{dim}, LFM40k~\cite{Xu2021VirtualMS}\\
 & TIMI-Net~\cite{Liu2021TripartiteIM} &ICCV&RGB-Trimap &&object& Parallel three-stream CNN & \makecell[c]{DIM-481~\cite{dim}, HATT-646~\cite{hatt},\\Human-2k~\cite{Liu2021TripartiteIM}} & \makecell[c]{DIM-481~\cite{dim}, HATT-646~\cite{hatt},\\Human-2k~\cite{Liu2021TripartiteIM}}\\
 & FGI~\cite{Cheng2021DeepIM} &BMVC& RGB-Flexible& & object & One-stage CNN & DIM-481~\cite{dim} &DIM-481~\cite{dim} \\
 & HEMatting~\cite{Zhong2021HighlyEN} &BMVC&RGB &\checkmark& object & Sequential two-stage CNN & DIM-481~\cite{dim} & DIM-481~\cite{dim}\\
 \hline
 \multirow{12}{*}{\rotatebox{90}{2022}} & RMat~\cite{dai2022boosting} &CVPR&RGB-Trimap & & object & Parallel two-stream CNN/Transformer & DIM-481~\cite{dim}& DIM-481~\cite{dim}, AIM-500~\cite{aim}\\
 & DIIM~\cite{Ding2022DeepII} &TIP& RGB-Click& & object & One-stage CNN & DIM-481~\cite{dim}, FI2O~\cite{Ding2022DeepII} & DIM-481~\cite{dim}\\
 & UGDMatting~\cite{Fang2022UserGuidedDH} &TIP&RGB-Flexible && human& Parallel two-stream CNN & UGD-12k~\cite{Fang2022UserGuidedDH}, HATT-646~\cite{hatt} & UGD-12k~\cite{Fang2022UserGuidedDH}, HATT-646~\cite{hatt}\\
 & matting-GP~\cite{Zheng2022ImageMW} &TNNLS& RGB-Trimap& & object &One-stage CNN & DIM-481~\cite{dim}, real-set2~\cite{Zheng2022ImageMW}& DIM-481~\cite{dim}, real-set2~\cite{Zheng2022ImageMW}\\\
 & P3M-ViTAE~\cite{p3mj} & IJCV & RGB &\checkmark  & human & Parallel two-stream CNN/Transformer & P3M-10k~\cite{p3m}&P3M-10k~\cite{p3m} \\
 & SPG-IM~\cite{Xu2022SituationalPG} & MM & RGB &\checkmark& object & Sequential two-stage CNN & \makecell[c]{DIM-481~\cite{dim}, Multi-Object-1K~\cite{Xu2022SituationalPG}\\Human-2k~\cite{tripathi2019learning}, HATT-646~\cite{hatt}} & \makecell[c]{DIM-481~\cite{dim}, Multi-Object-1K~\cite{Xu2022SituationalPG}\\Human-2k~\cite{tripathi2019learning}, HATT-646~\cite{hatt}}\\
 & HIM~\cite{Sun2022HumanIM} &CVPR& RGB&\checkmark&human & Sequential two-stage CNN & HIM2k~\cite{Sun2022HumanIM} & \makecell[c]{HIM2k~\cite{Sun2022HumanIM}, RWP636~\cite{mgmatting}\\SPD~\cite{sps}, COCO~\cite{coco}}\\
 & MatteFormer~\cite{Park2022MatteFormerTI} &CVPR&RGB-Trimap & & object& One-stage CNN &DIM-481~\cite{dim} & DIM-481~\cite{dim}\\
 & RIM~\cite{rim} & CVPR & RGB-Language& & object & Parallel two-stream CNN/Transformer & RefMatte~\cite{rim}&RefMatte~\cite{rim} \\
 & TransMatting~\cite{cai2022transmatting} & ECCV & RGB-Trimap &  & transparent& One-stage CNN/Transformer &Trans-460~\cite{cai2022transmatting} & Trans-460~\cite{cai2022transmatting} \\
 \hline
\end{tabular}}
\end{threeparttable}
\vspace{-10pt}
\end{table*}

\subsection{Traditional Solutions}

Prior to the advent of deep learning, researchers have made significant efforts to tackle image matting problems using traditional solutions. In the initial stages, the input image usually has a blue, green, or constant-color background, which is known as \textit{blue screen matting} \cite{Smith1996BlueSM,1571698600209592960}. It can reduce the difficulty of the problem and make it more tractable. Furthermore, \textit{triangular matting} is derived \cite{levin2008spectral,wang2007optimized} to assist in generating the ground-truth alpha mattes. To expand to natural images with complex backgrounds or even videos, additional information was utilized to constrain the problem, including trimap~\cite{chen2013knn,wang2007optimized,Sun2004PoissonM}, scribble~\cite{Levin2006ACS,zheng2008fuzzymatte,guan2006easy}, flash or non-flash image pairs~\cite{Sun2006FlashM}, camera arrays~\cite{Joshi2006NaturalVM}, and multiple synchronized video streams~\cite{McGuire2005DefocusVM}.

The traditional solutions for matting can be categorized into three types. Firstly, color sampling-based methods rely on the strong correlation between nearby image pixels to sample from known foreground or background colors and apply them to unknown pixels~\cite{ruzon2000alpha,Chuang2001ABA,Yang2003ImprovedFG}. Secondly, affinity-based methods calculate the affinity matrix to characterize the similarity between neighboring pixels and propagate alpha values from known areas to unknown areas accordingly~\cite{Sun2004PoissonM,Rother2004GrabCutIF,Bai2007AGF,Yang2003ImprovedFG,zheng2008fuzzymatte,Li2019ASO,levin2008spectral}. Thirdly, the combination of both sampling-based and affinity-based methods is adopted for optimization to achieve more robust solutions~\cite{wang2005iterative,Weiss2001OnTO,wang2007optimized,Szeliski2006LocallyAH}. Although these methods have shown significant improvements in predicting results through comprehensive design, their representation abilities are limited by low-level color or structure features, and they struggle to distinguish foreground details from complex natural backgrounds. Additionally, since most of these methods require manually labeled auxiliary inputs, the results are usually very sensitive to the size of the unknown region and fuzzy boundaries.


\begin{figure*}[t]
    \centering
    \includegraphics[width=\linewidth]{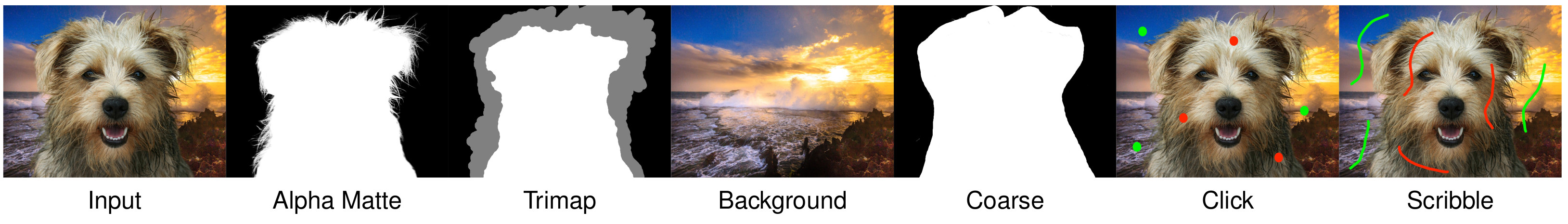}
    \caption{Illustration of a typical input image along with its alpha matte and various auxiliary inputs such as trimap, background, coarse map, user clicks, scribbles, and a text description that are commonly employed in auxiliary input-based image matting methods. The text description of the image can be \texttt{the cute smiling brown dog in the middle of the image}. In the click and scribble inputs, red lines or dots denote the pure foreground, while green lines or dots indicate the pure background. The foreground and background images used in this illustration are sourced from the AM-2k and BG-20k datasets~\cite{gfm}. We suggest enlarging the image for a more detailed view.
    }
    \label{fig:inputs}
\end{figure*}

\section{Auxiliary Input-based Image Matting}
\label{sec:auxiliary}
\begin{figure}[htbp]
\centering
\captionsetup[subfloat]{labelformat=empty,justification=centering}
\subfloat[]{\includegraphics[width=.98\linewidth]{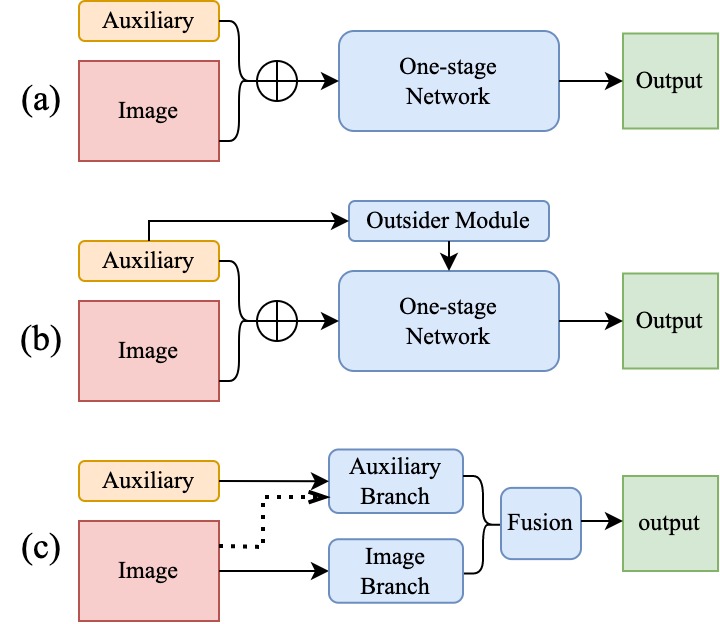}}
\caption{The architectures of the auxiliary input-based matting methods can be categorized into one of three types: (a) a one-stage model; (b) a one-stage model with an outsider module that processes auxiliary information; or (c) a two- or multi-stream model that processes the auxiliary input and the input image in separated streams.
}
\label{fig:architecture_auxiliary}
\vspace{-10pt}
\end{figure}

In this section, we provide a comprehensive review of the different types of auxiliary input-based matting models that utilize various input modalities such as trimap, scribble, background, coarse map, user interactive click, and text descriptions. An example of the input image and different auxiliary inputs is shown in Figure~\ref{fig:inputs}. Additionally, we summarize the architectures of the auxiliary input-based matting methods in Figure~\ref{fig:architecture_auxiliary}, which can be categorized into three types: 1) a one-stage model that processes the concatenation of the input image and auxiliary input; 2) a one-stage model that processes the concatenation of the input image and auxiliary input, and leverages the information of the auxiliary input with an outsider module; and 3) two- or multi-stream model that processes input image, auxiliary input, or their combination separately, and then passes through a fusion model to generate the final output. We discuss the details of these methods, along with their advantages, challenges, and comparisons with each other in the following part.

\subsection{Trimap-based Auxiliary Input}
After transitioning from blue screen matting to natural image matting, researchers began using the auxiliary input of a \textit{trimap} \cite{berman2000method,ruzon2000alpha}. It is a three-class map that indicates the definite foreground, definite background, and unknown region, as shown in Figure~\ref{fig:inputs}. A trimap is typically defined by the user or generated by off-the-shelf segmentation models~\cite{He2017MaskR} with dilation operations~\cite{0Comprehensive}. Although it significantly reduces the difficulty of the problem, it requires manual effort, which remains a challenge to achieve a good trade-off. With the advent of deep learning, advanced solutions utilize one of the three architectures shown in Figure~\ref{fig:architecture_auxiliary}.

\subsubsection{One-stage Architecture} 
The initial attempts of deep learning-based approaches to natural image matting with trimap input typically adopt a simple one-stage architecture. For example, DIM~\cite{dim} first solves the problem by concatenating the image and trimap as input and processing it through a fully convolutional network~\cite{Shelhamer2014FullyCN} with a U-net skip connection~\cite{unet} based on VGG-16~\cite{Simonyan2014VeryDC} to predict the output. FBA~\cite{Forte2020FBA} utilizes a similar structure based on ResNet-50~\cite{he2016deep} as the backbone but predicts seven channels instead of one, including one for the alpha matte, three for the foreground, and three for the background. AlphaGAN~\cite{bmvcLutzAS18} and MatteFormer~\cite{Park2022MatteFormerTI} adopt a similar one-stage architecture but with a generative adversarial network (GAN)~\cite{Karras2018ASG} and Swin Transformer~\cite{Liu2021SwinTH}, respectively. SmartScribbles~\cite{Yang2020SmartSF} oversamples the image into superpixels and rectangular regions and automatically selects the most informative ones for users to draw scribble, then propagates it through a two-phase propagation. TransMatting~\cite{cai2022transmatting} presents a trimap-guided transformer block~\cite{Liu2021SwinTH} between the ResNet-34 encoder and decoder to maintain the contexture of transparent objects.

\subsubsection{One-stage Architecture with Outsider Module} 
While the one-stage architecture used in some deep learning-based methods is simple and intuitive, it has limitations in exploiting auxiliary information. To address this issue, some methods~\cite{Wang2018DeepPB,lu2019indices,gca,sim,Zheng2022ImageMW} have introduced outsider modules that specifically process auxiliary information or the concatenation of auxiliary and input. For example, DMPN~\cite{Wang2018DeepPB} uses a module to extract low-level features and then applies a matte propagation module as a refiner. IndexNet~\cite{lu2019indices} proposes a module that dynamically predicts indices for individual local regions, which are then used to preserve details through downsampling and upsampling stages. GCA~\cite{gca} designs a module to directly propagate high-level opacity information globally based on learned low-level affinity. SIM~\cite{sim} uses a patch-based classifier to incorporate semantic classification of matting regions by extending the conventional trimap to a semantic one. Matting-GP~\cite{Zheng2022ImageMW} provides a Gaussian process to encapsulate the expressive power of deep architecture and reduce computation complexity.

\subsubsection{Multi-stream Architecture}
Various methods in recent years~\cite{Tang2019LearningBasedSF,cai2019disentangled,hou2019context,Li2020HierarchicalOP,Yu_Xu_Huang_Zhou_Shi_2021,Liu2021TowardsEF,Liu2021LongRangeFP,Liu2021TripartiteIM,dai2022boosting} have acknowledged the advantages of processing diverse levels of multi-task information through multi-stream encoders and subsequently fusing it via a shared decoder. For instance, SampleNet~\cite{Tang2019LearningBasedSF} estimates same-level foreground and background via a two-stream sample-selection process before predicting the opacity of the whole image. TIMI-Net~\cite{Liu2021TripartiteIM}, on the other hand, employs a three-branch encoder to supplement the neglected coordination between RGB space and trimap space, along with additional interactions. Other approaches strive to leverage global context information and local details simultaneously for sub-task aware propagation before the fusion network, such as the adapted trimap and alpha in AdaMatting~\cite{cai2019disentangled}, the foreground and alpha in Context-aware~\cite{hou2019context}, the image appearance and alpha opacity in HOP~\cite{Li2020HierarchicalOP}, the image patch with trimap and context patch with trimap in HDMatt~\cite{Yu_Xu_Huang_Zhou_Shi_2021}, the semantic path containing an image with trimap and a textural path containing an image with noisy trimap in FDMatting~\cite{Liu2021TowardsEF}, the downsampled images' context and surrounding image patches' features in LFPNet~\cite{Liu2021LongRangeFP}, and the transformer-based context and convolutional-based details in RMat~\cite{dai2022boosting}. These solutions lead to a deeper understanding of different information levels and a learnable fusion model for better collaboration.

\subsection{Pre-defined Auxiliary Input}
In addition to the commonly used trimap, some methods employ pre-defined images, such as backgrounds~\cite{backgroundmatting,backgroundmattingv2}, or coarse maps~\cite{Cho2016NaturalIM,mgmatting} as auxiliary inputs.

\subsubsection{Background-based Matting Model} 
Due to the time-consuming nature of obtaining trimaps, some researchers have proposed alternative methods for obtaining auxiliary inputs. For instance, BM~\cite{backgroundmatting} and BMV2~\cite{backgroundmattingv2} require an additional photo of the background, which is captured by the user beforehand, as shown in Figure~\ref{fig:inputs}. However, these methods are often limited to human matting targets. In BM~\cite{backgroundmatting}, a soft segmentation is first computed and then passed through a network~\cite{mao2017least} to estimate the foreground and alpha matte with an adversarial loss. On the other hand, BMV2~\cite{backgroundmattingv2} provides a base network to compute a low-resolution result, which is then refined by a second network that operates at high-resolution on patches.

\subsubsection{Coarse-based Matting Model}
Although background-based matting provides good results, it has limitations in certain scenarios and requires some level of foresight. Consequently, some researchers have explored the use of coarse segmentation maps~\cite{Cho2016NaturalIM,mgmatting} as inputs instead, as shown in Figure~\ref{fig:inputs}. For instance, DCNN~\cite{Cho2016NaturalIM} produces auxiliary coarse maps using CF~\cite{Levin2006ACS} and KNN~\cite{chen2013knn}, which are then combined with the original image and fed into a one-stage CNN network. A2U~\cite{dai2021learning} proposes to leverage second-order features to formulate an affinity-aware upsampling block, which replaces the normal block in a ResNet34-based one-stage CNN structure. MGMatting~\cite{mgmatting} adopts an encoder-decoder architecture with the image and coarse map as inputs, using a self-guidance mechanism to progressively refine uncertain regions.

\subsection{User-interactive Auxiliary Input}
To address the limitations of pre-defined auxiliary inputs, researchers have explored the use of user-interactive auxiliary inputs. These methods~\cite {NEURIPS2018_653ac11c,Wei2020ImprovedIM,Ding2022DeepII,Cheng2021DeepIM,Fang2022UserGuidedDH} allow users to interactively provide auxiliary inputs that can dynamically refine the results.

\subsubsection{Click-based Matting Model}
Several works~\cite{NEURIPS2018_653ac11c, Wei2020ImprovedIM, Ding2022DeepII} incorporate user click, as shown in Figure~\ref{fig:inputs}, as an auxiliary input. For example, ActiveMatting~\cite{NEURIPS2018_653ac11c} proposes a recurrent reinforcement learning framework~\cite{Hochreiter1997LongSM,Williams1992SimpleSG} that involves human interaction through clicks. InteractiveMatting~\cite{Wei2020ImprovedIM} uses a two-decoder network to take RGB image and click as input and refines the result using an uncertainty-guided local refinement module. DIIM~\cite{Ding2022DeepII} transforms user clicks into a distance map, concatenates it with the original image, and passes it through an encoder-decoder network and a full-resolution extraction module to predict the alpha matte.

\subsubsection{Flexible User Input-based Matting Model} 
Other approaches~\cite{Cheng2021DeepIM,Fang2022UserGuidedDH,Yang2020SmartSF} in the field choose more flexible inputs as auxiliary signals, such as scribble. For instance, FGI~\cite{Cheng2021DeepIM} employs an encoder-decoder structure that gradually reduces the area of the uncertain region in the auxiliary input, thereby improving the prediction of the final alpha matte. UGDMatting~\cite{Fang2022UserGuidedDH} uses a double-encoder and double-decoder to extract image features, propagate user interaction data, and generate foreground and background before a fusion model predicts the final result. SmartScribbles~\cite{Yang2020SmartSF} guides users to draw only a few scribbles to achieve high-quality matting results.

\subsection{Text-based Auxiliary Input} 
In recent research, there has been a trend to explore the integration of other input modalities with the input image to enable a more controllable matting procedure. For example, text descriptions have been used to provide flexible auxiliary information. One representative work is RIM~\cite{rim}, which utilizes a CLIP-based~\cite{Radford2021LearningTV} two-branch encoder-decoder framework to extract text-driven semantic features from the original image while preserving local matting details. Additionally, RIM has established RefMatte, a large-scale benchmark dataset with massive amounts of paired image and text descriptions, which will be discussed in detail in Section~\ref{sec:dataset}.

\subsection{Summary} 
Data-driven methods that utilize auxiliary inputs have become prevalent in the era of deep learning, as they can provide more precise results than fully automatic approaches. In recent years, various forms of auxiliary inputs have been explored to provide different benefits, such as trimap, background, coarse map, click map, scribble, and even text description. With the development of this field, the form of auxiliary input has become increasingly flexible and less time-consuming, allowing for more interactive and intuitive matting procedures.

As previously discussed, the methodologies for auxiliary input-based image matting can be broadly classified into three types. The first type is the one-stage network, which directly predicts the alpha matte from the input image and the auxiliary input. The second type is the one-stage network with a specially designed external module that processes the auxiliary input before fusing it with the image features to obtain the final alpha matte. The third type is the multi-stream model, which processes the input and the auxiliary input at different levels before fusing them to generate the final output. Each of these methods improves upon the previous one by utilizing the features from the auxiliary input and performing subtask-aware global-local collaboration.

Despite their success, these methods may have two potential limitations. Firstly, they still require varying degrees of manual effort, which may not be feasible for automatic industrial applications. Additionally, certain inputs such as background images may require foresight that is not available in many practical scenarios. Secondly, most of these methods exhibit high sensitivity to specific auxiliary inputs, such as the size of the transition area in trimap, the accuracy of the coarse map, the density of the clicks, and the shape of strokes, making the development of robust methods a significant challenge.

\section{Automatic Image Matting}
\label{sec:automatic}

\begin{figure}[htbp]
\centering
\captionsetup[subfloat]{labelformat=empty,justification=centering}
\subfloat[]{\includegraphics[width=.98\linewidth]{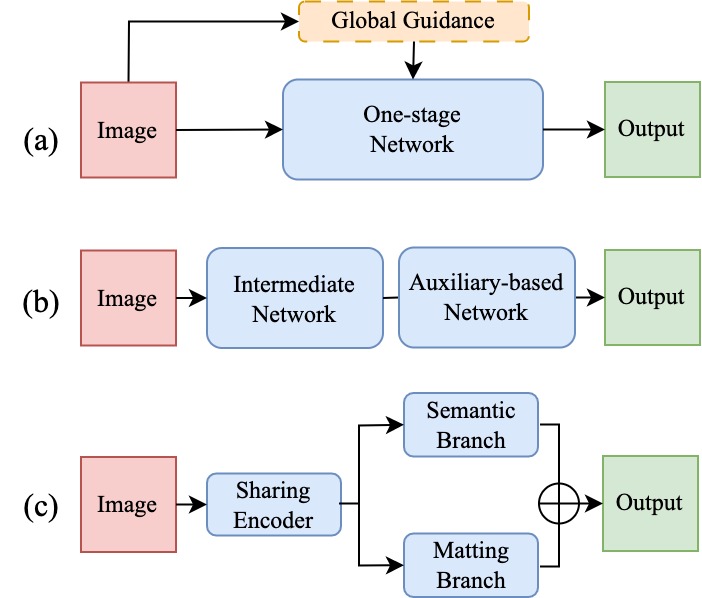}}
\caption{
The architectures of the automatic matting methods can be classified into one of three types: (a) a one-stage model with an optional global guidance module; (b) a sequential two-step model that predicts intermediate and final outputs; or (c) an encoder-sharing network with two or more separate decoders.
}
\label{fig:architecture_automatic}
\vspace{-10pt}
\end{figure}

To overcome the limitations of auxiliary input-based matting methods and render them applicable to real-world industrial settings, scholars have put forward automatic matting methods that predict the alpha matte from the input image without relying on any auxiliary input. In this section, we review these methods, focusing on their architectures as illustrated in Figure~\ref{fig:architecture_automatic}, as well as their matting targets.

\subsection{One-stage Matting Model}
Several studies, including HATT~\cite{hatt}, MODNet~\cite{modnet}, and CasDGR~\cite{Yu2021CascadeIM}, propose a one-stage model with an optional guidance module, as illustrated in Figure~\ref{fig:architecture_automatic}(a). HATT employs a ResNeXt~\cite{Xie2016AggregatedRT} network that utilizes skip-connections, along with a global guidance module that incorporates channel-wise and spatial-wise attention mechanisms, to predict alpha matte directly from the input image. Similarly, MODNet employs a MobileNetV2~\cite{sandler2018mobilenetv2} CNN as its backbone network and incorporates several convolutional layers as its global guidance. CasDGR utilizes graph neural networks (GNNs)~\cite{Zhang2022DynamicGM,Yu2020RepresentativeGN} to optimize its one-stage network, producing the output from coarse to fine.

\subsection{Sequential Matting Model}
An alternative solution to automatic matting is a sequential two-step architecture, depicted in Figure~\ref{fig:architecture_automatic}(b). In this method, the first network is utilized to generate an intermediate representation, such as a trimap, a coarse map, or a combination of the foreground, background, and unknown regions. Its goal is to alleviate the difficulty of the problem by simplifying the second stage, and transforming it into a straightforward auxiliary input-based matting problem. This methodology is widely employed and has been adopted in numerous studies.

\subsubsection{Trimap as the Intermediate}
The most straightforward intermediate representation is the trimap. DAPM~\cite{dapm} and SHM~\cite{shm} predict it through the intermediate network, \eg, FCN-8s~\cite{Shelhamer2014FullyCN} in DAPM and PSPNet-50~\cite{Zhao2016PyramidSP} in SHM. These two methods are both tailored for human matting.

\subsubsection{Coarse Map as the Intermediate}
Another commonly used intermediate representation is the coarse map, which can be generated using an off-the-shelf segmentation network such as MaskRCNN~\cite{He2017MaskR} as in HIM~\cite{Sun2022HumanIM}, or a carefully designed network as in FDM~\cite{Zhu2017FastDM}, SHMC~\cite{shmc}, and SPG-IM~\cite{Xu2022SituationalPG}. Specifically, FDM~\cite{Zhu2017FastDM} uses densely connected blocks~\cite{huang2017densely} with dilated convolution~\cite{chen2017deeplab} as the intermediate network, while SHMC~\cite{shmc} utilizes an encoder-decoder CNN with skip connections. SPG-IM~\cite{Xu2022SituationalPG} adopts a ResNet-50~\cite{he2016deep} captioning encoder, a transformer-based~\cite{NIPS2017_3f5ee243} text decoder, and an Atrous Spatial Pyramid Pooling (ASPP)~\cite{chen2017deeplab} based encoder-decoder. As for the auxiliary-based network in the second stage, HIM~\cite{Sun2022HumanIM} utilizes the one provided in MGMatting~\cite{mgmatting}, while FDM~\cite{Zhu2017FastDM} devises a feature block with guided filter~\cite{He2010GuidedIF}. SHMC~\cite{shmc} chooses an encoder-decoder with skip-connections, and SPG-IM~\cite{Xu2022SituationalPG} leverages a ResNet-50 encoder-decoder.

\subsubsection{Other Intermediate}
Some other works use different types of intermediate representations, such as foreground and background~\cite{lf}, unknown regions~\cite{Zhong2021HighlyEN}, depth maps and heatmaps~\cite{Xu2021VirtualMS}, albedo maps, shading maps, and specular maps~\cite{Chen2018TOMNetLT}. In LF~\cite{lf}, the foreground and background probability maps are obtained using a DenseNet-201~\cite{huang2017densely}-based encoder and then fed into an FCN fusion network for prediction. HEMatting~\cite{Zhong2021HighlyEN} employs an OctConv~\cite{Chen2019DropAO}-based encoder-decoder network to produce 3-channel classification logits and another one for the final output. VMFM~\cite{Xu2021VirtualMS} generates a depth map through DenseDepth~\cite{Alhashim2018HighQM}, a human segmentation map through Mask-RCNN~\cite{He2017MaskR}, and a human-object paired heatmap following the work~\cite{Wang2020LearningHI} as the intermediate representations. It then produces alpha matte using a dual network and a complementary learning module to estimate the deviation probability map for predicting the final output. TOM-Net~\cite{Chen2018TOMNetLT} uses a mirror-link CNN~\cite{Shi2016LearningNO} to predict an albedo map, a shading map, and a specular map, which are then concatenated with the input image and passed through a residual network for obtaining the final result. SSS~\cite{sss} extracts high-level semantic features from DeepLab-ResNet-101~\cite{chen2017deeplab} and proposes a graph structure to predict soft labels using the Laplacian matrix from embedding texture and color features.

\subsection{Encoder Sharing Model}
Although the previous two architectures offer an intuitive design for automatic matting, recent works~\cite{gfm, aim, p3m, p3mj} explicitly decompose the task into several sub-tasks, learn mutual features through sharing an encoder, and formulate each sub-task as a separate decoder branch, as shown in Figure~\ref{fig:architecture_automatic}(c). Each decoder output should be supervised by the relevant ground truth to ensure that each sub-task is optimized properly. Finally, a hard fusion mechanism is used to produce the final result from the outputs of all the decoders. It typically has two decoders for global semantic segmentation and local matting, respectively. In the semantic decoder, the output can be foreground and background masks~\cite{gfm}, or trimap~\cite{gfm, aim, p3m, p3mj}. GFM~\cite{gfm} uses DenseNet-121~\cite{huang2017densely}, ResNet-34~\cite{he2016deep}, or ResNet-101 as the backbone of the shared encoder, a pyramid pooling module (PPM)~\cite{Zhao2016PyramidSP,Liu2019ASP} in the global decoder and a bridge block~\cite{Qin_2019_CVPR} in the local decoder. AIM-Net~\cite{aim} devises a spatial attention module to guide the matting decoder by leveraging the learned semantic features from the semantic decoder to focus on extracting details within the transition area. P3M-Net~\cite{p3m} designs a tripartite-feature integration, a shallow bipartite-feature integration, and a deep bipartite-feature integration to model interactions among the shared encoder and two decoders. Specifically, P3M-ViTAE~\cite{p3mj} adopts the ViTAE transformer~\cite{vitae,zhang2023vitaev2} as the shared encoder to better represent the long-range global context and local detail features.

\subsection{Matting Target}
Automatic matting methods, which are designed without any auxiliary input, are prone to training data bias. As a result, most of these methods are limited to specific matting targets, such as human~\cite{modnet}, animal~\cite{gfm}, and transparent objects~\cite{Chen2018TOMNetLT}. In this regard, we provide a detailed discussion and comparison of the existing methods. Several methods, such as DAPM~\cite{dapm}, SHM~\cite{shm}, CasDGR~\cite{Yu2021CascadeIM}, FDM~\cite{Zhu2017FastDM}, SHMC~\cite{shmc}, and MODNet~\cite{modnet}, focus on the most prevalent matting target, \ie, \textit{human}. Some methods, including VMFM~\cite{Xu2021VirtualMS} and SPG-IM~\cite{Xu2022SituationalPG}, focus on human-object interaction. HIM~\cite{Sun2022HumanIM} is designed for predicting the alpha matte of human instances. P3M-Net~\cite{p3m} and P3M-ViTAE~\cite{p3mj} produce an alpha matte for privacy-preserving human images. GFM~\cite{gfm} is designed to process both human and animal images, while TOM-Net~\cite{Chen2018TOMNetLT} is trained to learn the opacity map of transparent objects. Other methods, such as HATT~\cite{hatt}, LF~\cite{lf}, and HEMatting~\cite{Zhong2021HighlyEN}, work on normal types of objects. AIM-Net~\cite{aim} further divides them into salient opaque foregrounds, salient transparent/meticulous foregrounds, and non-salient foregrounds, with a carefully designed generalized trimap for each type.

\begin{table*}[t]
\centering
\caption{Summary of image matting datasets, categorized as the synthetic image-based benchmark, natural image-based benchmark, and test sets. The datasets are ordered based on their release date and are described in terms of publication venue, source modality, label type, naturalness, matting target, resolution, annotation method, number of training and test samples, and availability. It should be noted that the size of the datasets is calculated based on the number of distinguished foregrounds, except for TOM~\cite{Chen2018TOMNetLT} and RefMatte~\cite{rim}, which have pre-defined composite rules.
}
\vspace{-5pt}
\label{tab:matting_datasets}
\begin{threeparttable}
\resizebox{0.98\textwidth}{!}{
\setlength\tabcolsep{6pt}
\renewcommand\arraystretch{1.15}
\begin{tabular}{|r|c||c|c|c|c|c|c|c|c|c|c|}
\hline
\rowcolor{figure_head}
Method &Pub. & Modality & Label & Natural & Target & Resolution & Annotation & \#Train  & \#Test  & Public \\
\hline
\hline
 DIM-481~\cite{dim} & CVPR'17 & Image, Video frame & alpha & & object & $1,298\times1,083$  & Manually & 431 & 50 & \checkmark \\
 TOM~\cite{Chen2018TOMNetLT} & CVPR'18 & 3D model& alpha, refractive flow & & transparent & - & POV-Ray~\cite{povray} & 178,000  & 876 & \checkmark\\
 LF-257~\cite{lf} & CVPR'19 & Image & alpha & & human & $553\times756$  & Manually & 228 & 29 & \checkmark\\
 HATT-646~\cite{hatt} & CVPR'20 & Image & alpha &\ & object &  $1,573\times1,731$ & Manually & 596 & 50 & \checkmark\\
 PhotoMatte13k~\cite{backgroundmattingv2} & CVPR'20 & Image& alpha & & human & - & Manually & 13,665 & - & \\
 SIM~\cite{sim} & CVPR'21 & Image& alpha & & object & $2,194\times1,950$  & Manually & 348 & 50 & \checkmark\\
 Human-2k~\cite{Liu2021TripartiteIM} & ICCV'21 & Image& alpha & & human & $2,112\times2,075$   & Manually & 2,000 & 100 & \checkmark\\
 Trans-460~\cite{cai2022transmatting} & ECCV'22 & Image& alpha & & transparent &  $3,766\times3,820$   & Manually & 410 & 50 & \checkmark\\ 
 HIM2k~\cite{Sun2022HumanIM} & CVPR'22 & Image& alpha & & human &  $1,823\times1,424$   & Manually & 1500 & 500 & \checkmark\\ 
 RefMatte~\cite{rim} & CVPR'23 & Image& alpha, language & & object & $1,543\times1,162$  & Manually & 45,000  & 2,500 & \checkmark\\
 
 \hline
 AlphaMatting~\cite{rhemann2009perceptually} & CVPR'09 & Image& alpha &\checkmark & object &  $3,056\times2,340$ & Manually & 27 & 8 & \checkmark\\
 DAPM-2k~\cite{dapm} & ECCV'16 & Image& alpha &\checkmark & human & $600\times800$  & KNN~\cite{chen2013knn}, CF~\cite{Levin2006ACS} & 1700 & 300 & \checkmark\\
 SHM-35k~\cite{shm} & MM'18 & Image& alpha &\checkmark & human &  - & Manually &52,511  & 1,400 & \\
 SHMC-10k~\cite{shmc} & CVPR'20 & Image& alpha, coarse map &\checkmark & human &  - & Manually &9,324 & 125 & \\
 P3M-10k~\cite{p3m} & MM'21 & Image& alpha &\checkmark & human & $1,349\times1,321$  & Manually & 9,421 & 1,000 & \checkmark\\
 AM-2k~\cite{gfm} & IJCV'22 & Image& alpha &\checkmark & animal & $1,471\times1,195$  & Manually & 1,800 & 200 & \checkmark\\
Multi-Object-1K~\cite{Xu2022SituationalPG} & MM'22 & Image& alpha &\checkmark & human-object &  - & Manually & 1000 & 200 & \\
UGD-12k~\cite{Fang2022UserGuidedDH} & TIP'22 & Image& alpha &\checkmark & human &  $356\times317$ & Manually & 12,066 & 700 & \checkmark\\

\hline
\hline
PhotoMatte85~\cite{backgroundmattingv2} & CVPR'20 & Image& alpha & & human &  $2,304\times3,456$  & Manually & - & 85 & \checkmark\\
AIM-500~\cite{aim} & IJCAI'21 & Image& alpha &\checkmark & object & $1,397\times1,260$  & Manually & - & 500 & \checkmark\\
RWP-636~\cite{mgmatting} & CVPR'21 & Image& alpha, detail map &\checkmark & human & $1,038\times1,327$  & Manually & - & 636 & \checkmark\\
PPM-100~\cite{modnet} & AAAI'22 & Image& alpha &\checkmark & human &  $2,997\times2,875$   & Manually & - & 100 & \checkmark\\ 
 \hline
\end{tabular}}
\end{threeparttable}
\vspace{-10pt}
\end{table*}

\subsection{Summary}

Automatic methods have been developed to reduce the need for time-consuming manual labelling of auxiliary inputs. These methods can be categorized into three types of architectures: 1) a one-stage network with an optional global guidance module, 2) a sequential stage-wise network that predicts intermediate representation and produces the alpha matte sequentially, and 3) an encoder-sharing networks with separate decoders that generate sub-tasks' outputs before a final fusion block. Various intermediate representations have been investigated, including trimap, coarse map, foreground, background, and unknown regions, depth map, heatmap, shading map, and others. However, due to the high demand for training data, these methods are prone to biases towards specific matting targets, such as humans, human-object interaction, animals, transparent objects, salient objects, and non-salient objects.

Despite the advantages of automatic matting methods for practical industrial applications, there are still some limitations that need to be addressed. Firstly, without auxiliary inputs, these methods tend to extract all salient or centralized foregrounds, which can result in less controllable outcomes that require significant manual effort to improve. Secondly, due to the bias towards training data, it is important to consider how to generalize these methods to new categories and unseen data. Lastly, to be applied in industrial settings, it is crucial to develop models that are lightweight and computationally efficient, capable of operating in real-time.

\section{Image Matting Datasets}
\label{sec:dataset}

Deep learning-based methods rely heavily on high-quality training and testing datasets~\cite{vanderPloeg2014ModernMT}, and this holds true for the task of image matting as well. This section provides a comprehensive review of image matting datasets, spanning from traditional methods to contemporary ones. We categorize the datasets into two types: synthetic image-based datasets and natural image-based datasets. Table~\ref{tab:matting_datasets} provides a summary of these datasets, and we discuss their details in the following subsections.

\subsection{Synthetic Matting Datasets}

Because it is time-consuming to manually label meticulous alpha mattes of foregrounds, many matting datasets~\cite{dim,lf,hatt,sim,rim} consist of synthetic images created by extracting the foregrounds from simple or plain backgrounds and compositing them with complex backgrounds using alpha blending~\cite{Levin2006ACS}. DCNN~\cite{Cho2016NaturalIM} first introduces a synthetic dataset by compositing 27 foregrounds from AlphaMatting~\cite{rhemann2009perceptually} with numerous backgrounds. DIM~\cite{dim} introduces 481 foregrounds, with 27 from AlphaMatting~\cite{rhemann2009perceptually}, some frames from video datasets~\cite{Shahrian2014TemporallyCA}, and images from plain backgrounds. The alpha mattes are manually generated by Photoshop~\cite{adobephotoshop}, and the backgrounds used in composition are from MS COCO~\cite{coco} and Pascal VOC~\cite{Everingham2010ThePV}. TOM~\cite{Chen2018TOMNetLT} renders multiple 3D models using POV-ray~\cite{povray} to create large amounts of synthetic transparent datasets. LF~\cite{lf} contains 257 human images from the internet with carefully annotated alpha mattes. HATT~\cite{hatt} provides 646 distinct foregrounds of multiple categories to improve the diversity of the matting datasets. PhotoMatte~\cite{backgroundmattingv2} has 13k synthetic high-resolution human foregrounds, but only 85 test images are publicly available. SIM~\cite{sim} collects 398 images from 20 classes that are widely used in matting scenarios and annotates the alpha mattes using Photoshop. Human-2k~\cite{Liu2021TripartiteIM} is a high-resolution human dataset containing 2,100 synthetic images. RIM~\cite{rim} establishes RefMatte containing 13,187 foregrounds from public synthetic matting datasets, along with their category labels, to form 45,000 training images and 2,500 test images with backgrounds from BG-20k~\cite{gfm}. RefMatte also provides text description labels to help develop models for referring image matting. Please note that the number of synthetic datasets shown in Table~\ref{tab:matting_datasets} is calculated on a per-foreground basis, and each foreground is usually blended with 20 to 100 backgrounds. Trans-460~\cite{cai2022transmatting} introduces 460 transparent object foregrounds to facilitate the study of transparent matting. HIM2k~\cite{Sun2022HumanIM} collects 2,000 human images with alpha mattes, of which 360 are natural images and the rest are synthetic. Please note that since TOM~\cite{Chen2018TOMNetLT} and RefMatte~\cite{rim} have predefined explicit composite rules for generating the training datasets, we report the actual amount of training data directly.

Although using synthetic datasets to train deep learning models for image matting enjoys the benefits of providing large amounts of training data with various complex backgrounds, it also results in a domain gap between composite images and natural images, due to the \textit{composition artifacts} that arise from discrepancies between the foregrounds and backgrounds~\cite{Blau2017ThePT,Swaminathan2008DigitalIF,gfm}. Numerous works have observed such phenomena~\cite{Chen2018TOMNetLT,Yang2020SmartSF,mgmatting,Liu2021TripartiteIM,Fang2022UserGuidedDH,Xu2022SituationalPG,Sun2022HumanIM,gfm,rim}, and some have focused on reducing these discrepancies and improving the generalization ability on real-world natural images~\cite{hou2019context,gfm,backgroundmatting,modnet,dai2022boosting}. For example, Context-Aware~\cite{hou2019context} applies re-JPEGing and Gaussian blur as data augmentation following the work~\cite{Khan2015ExposingDI,Huh2018FightingFN}. GFM~\cite{gfm} proposes a composition route called RSSN to address issues such as resolution, sharpness, noise, and illumination discrepancies. They also build a large-scale high-resolution dataset BG-20k that contains diverse background images. BM~\cite{backgroundmatting} describes a self-supervised scheme to learn from unlabelled real data and a discriminator to improve the result. MODNet~\cite{modnet} fine-tunes the model on unlabeled real data by using consistency between sub-objectives to reduce the domain gap. RMat~\cite{dai2022boosting} employs strong data augmentations, such as linear pixel-wise augmentation, non-linear pixel-wise augmentation, and region-wise augmentation, to mitigate the domain gap.

\subsection{Natural Matting Datasets}

To enhance the generalization capability of matting models to real-world images, some researchers have established natural image matting datasets, encompassing a diverse range of categories such as humans~\cite{p3m}, animals~\cite{gfm}, and objects~\cite{rhemann2009perceptually}. Table~\ref{tab:matting_datasets} provides a summary of these datasets, presenting their source modality, type of labels, matting target, average resolution, annotation method, as well as the number of training and testing images, and availability.

AlphaMatting~\cite{rhemann2009perceptually} captures high-quality ground truth mattes in a restricted studio environment using triangulation~\cite{Smith1996BlueSM} and augments the data~\cite{rhemann2008high} to yield a total of 35 high-resolution images. It also offers a complete online benchmark system with carefully designed error measures, which provides great value for trimap-based methods, especially in the traditional solutions era. However, with the emergence of data-driven deep learning methods, large-scale matting datasets have become essential. Therefore, DAPM~\cite{dapm} collects 2,000 human portrait images with ground truth labels generated by off-the-shelf methods KNN~\cite{chen2013knn} and CF~\cite{Levin2006ACS}. Subsequently, to enhance image resolution and label accuracy, researchers have established large-scale datasets with manually annotated alpha mattes focusing on humans~\cite{shm,shmc,p3m,Fang2022UserGuidedDH}, animals~\cite{gfm}, and human-objects~\cite{Xu2022SituationalPG}. However, many of these datasets are not public due to privacy or licensing issues, which impedes future research in the community. P3M-10k~\cite{p3m} addresses this issue by providing the first large-scale privacy-preserving portrait matting dataset with carefully annotated alpha mattes of 10,421 human images in various backgrounds. AM-2k~\cite{gfm} is the first animal matting dataset that contains 2,000 images from 20 categories. UGD~\cite{Fang2022UserGuidedDH} also provides a large number of public portrait images, although the resolution is somewhat limited for the matting task.

In addition to the aforementioned benchmark datasets, there exist several test-only real-world matting datasets that aim to facilitate the evaluation of matting models' generalization ability. For instance, AIM-500~\cite{aim} is the first natural image matting dataset that comprises 500 high-resolution images of different categories, along with alpha mattes labeled by professionals. RWP-636~\cite{mgmatting} introduces 636 real-world portrait images, where additional detail map labels are provided to indicate the most challenging detail regions. PPM-100~\cite{modnet} is another publicly available real-world human dataset, consisting of 100 images.

\subsection{Summary} 

High-quality datasets and accurate labels are essential for the success of deep learning-based image matting. However, creating such datasets is a significant challenge. To tackle this issue, researchers have established several valuable datasets that are publicly available. These datasets can be broadly classified into two categories: synthetic and composite datasets, each with its strengths and limitations.

Synthetic datasets require relatively little manual effort and can generate a large number of training images by easily composing each foreground with many various backgrounds. However, these synthetic images can exhibit composition artifacts that may introduce a significant domain gap when compared to real-world images. On the other hand, natural matting datasets do not suffer from such issues and the trained models can easily adapt to real-world images. However, they may be limited to specific matting targets and smaller scales due to the difficulty of accurately extracting alpha mattes from images in the wild. Additionally, the manual labelling process for generating ground truth alpha mattes may introduce errors or noise. Thus, developing a scalable approach for generating high-quality and large-scale image matting datasets remains a significant challenge.

\section{Performance Benchmarking}
\label{sec:performance}

In this section, we provide a comprehensive evaluation of representative matting methods. A thorough examination of the training objectives (refer to Table~\ref{tab:losses}) and evaluation metrics is presented. Additionally, a fair and rigorous benchmarking of both auxiliary-based and automatic models is carried out (refer to Table~\ref{tab:exp_results}), followed by an in-depth analysis of the results.

\begin{table}[htbp]
\centering
\caption{Summary of losses used in deep learning-based matting methods.}
\vspace{-5pt}
\label{tab:losses}
\resizebox{0.48\textwidth}{!}{
\setlength\tabcolsep{6pt}
\renewcommand\arraystretch{1.5}
\begin{tabular}{|l|c|c|}
\hline
& \makebox[0.2\textwidth][c]{Auxiliary} & \makebox[0.2\textwidth][c]{Automatic} \\
\hline
$\mathcal{L}_{alpha}$ & \makecell[l]{~\cite{dim},~\cite{Wang2018DeepPB},~\cite{bmvcLutzAS18},~\cite{dai2022boosting},~\cite{lu2019indices},\\\cite{cai2019disentangled},~\cite{hou2019context},~\cite{gca},~\cite{backgroundmatting},~\cite{Ding2022DeepII},\\
~\cite{Forte2020FBA}~\cite{Yu_Xu_Huang_Zhou_Shi_2021},~\cite{mgmatting},~\cite{Wei2020ImprovedIM},~\cite{backgroundmattingv2},\\~\cite{Liu2021TowardsEF},~\cite{sim},\cite{Liu2021LongRangeFP},~\cite{Liu2021TripartiteIM},~\cite{Cheng2021DeepIM},\\~\cite{Tang2019LearningBasedSF},~\cite{Li2020HierarchicalOP},~\cite{Fang2022UserGuidedDH},~\cite{Park2022MatteFormerTI},~\cite{cai2022transmatting}} &\makecell[l]{ ~\cite{dapm},~\cite{Zhu2017FastDM},~\cite{shm},~\cite{lf},~\cite{Sun2022HumanIM},\\~\cite{gfm},~\cite{modnet},~\cite{p3m},~\cite{Yu2021CascadeIM},~\cite{Xu2022SituationalPG},\\~\cite{Xu2021VirtualMS},~\cite{Zhong2021HighlyEN},~\cite{p3mj},~\cite{aim},~\cite{shmc}}\\
 \hline
 $\mathcal{L}_{comp}$ &\makecell[l]{~\cite{dim},~\cite{Wang2018DeepPB},~\cite{bmvcLutzAS18},~\cite{Tang2019LearningBasedSF},~\cite{lu2019indices},\\~\cite{backgroundmatting},~\cite{Forte2020FBA},~\cite{Yu_Xu_Huang_Zhou_Shi_2021},~\cite{mgmatting},~\cite{backgroundmattingv2},\\\cite{Liu2021LongRangeFP},~\cite{Cheng2021DeepIM},~\cite{dai2022boosting},~\cite{Park2022MatteFormerTI},~\cite{cai2022transmatting}} & \makecell[l]{~\cite{Zhu2017FastDM},~\cite{shm},~\cite{shmc},~\cite{gfm},~\cite{Xu2021VirtualMS},\\~\cite{p3m},~\cite{Yu2021CascadeIM},~\cite{aim},~\cite{modnet},~\cite{p3mj},\\~\cite{Sun2022HumanIM}}\\
\hline
$\mathcal{L}_{mse}$ & \makecell[l]{~\cite{Cho2016NaturalIM},~\cite{hou2019context},~\cite{Wei2020ImprovedIM},~\cite{Ding2022DeepII},~\cite{sim},\\~\cite{backgroundmattingv2},~\cite{Fang2022UserGuidedDH},~\cite{Zheng2022ImageMW},~\cite{NEURIPS2018_653ac11c}}&\makecell[l]{
~\cite{Chen2018TOMNetLT},~\cite{lf},~\cite{hatt},~\cite{modnet}}\\
\hline
$\mathcal{L}_{ce}$ &\makecell[l]{ ~\cite{cai2019disentangled},~\cite{Yang2020SmartSF},~\cite{sim},~\cite{Ding2022DeepII},~\cite{Fang2022UserGuidedDH}} & \makecell[l]{ ~\cite{Chen2018TOMNetLT},~\cite{shm},~\cite{lf},~\cite{gfm},~\cite{p3m},\\~\cite{aim},~\cite{Zhong2021HighlyEN},~\cite{p3mj}}\\
\hline
$\mathcal{L}_{lap}$ &\makecell[l]{
~\cite{hou2019context},~\cite{Forte2020FBA},~\cite{mgmatting},~\cite{dai2022boosting},~\cite{Park2022MatteFormerTI},\\~\cite{cai2022transmatting}} &\makecell[l]{~\cite{gfm},~\cite{p3m},~\cite{aim},~\cite{p3mj},~\cite{Sun2022HumanIM}}\\
\hline
$\mathcal{L}_{grad}$ &\makecell[l]{ ~\cite{Tang2019LearningBasedSF},~\cite{Forte2020FBA},~\cite{sim},~\cite{Cheng2021DeepIM},\\~\cite{dai2022boosting},~\cite{Fang2022UserGuidedDH}} & ~\cite{Yu2021CascadeIM},~\cite{Zhong2021HighlyEN}\\
\hline
$\mathcal{L}_{adv}$ & ~\cite{bmvcLutzAS18} & ~\cite{hatt},~\cite{Xu2021VirtualMS}\\
\hline
$\mathcal{L}_{multi}$ & ~\cite{cai2019disentangled} & - \\
\hline
$\mathcal{L}_{ssim}$& - & ~\cite{hatt}\\
 \hline
\end{tabular}
}
\vspace{-10pt}
\end{table}

\subsection{Training Objectives}

To enable the effective training of deep learning-based matting models, researchers have developed various loss functions. In this section, we outline the definitions of the most commonly used loss functions and present a summary of them in Table~\ref{tab:losses}, with a focus on their application in both auxiliary input-based models and automatic models.

\noindent\textbf{Alpha loss}
The alpha loss function, denoted as $\mathcal{L}_{alpha}$, is calculated as the absolute difference between the ground truth alpha matte, $\alpha_g$, and the predicted alpha matte, $\alpha_p$, on a per-pixel basis. To ensure the differentiability of the loss, a small value $\varepsilon$ is typically added~\cite{dim}, as demonstrated in Eq.~\eqref{equa:loss_alpha}. Here, $i$ denotes the index of the pixel and $\varepsilon$ is set to $10^{-6}$ for computational stability. The weight for each pixel, $W_{\alpha}^{i} \in \left[ 0,1 \right]$, is constantly set to 1 for automatic matting models, and for auxiliary-based methods, it is set to 1 in transition areas and 0 in the foreground and background regions. Some studies~\cite{dapm} have also utilized different weights for each pixel when calculating the loss.
\begin{equation}
\mathcal{L}_{alpha}= \frac{\sum_{i}\sqrt{\left ( \left ( \alpha_p^i-\alpha_g^i \right )\times W_{\alpha}^i \right )^{2}+\varepsilon^{2}}}{\sum_{i}W_{\alpha}^i}.
\label{equa:loss_alpha}
\end{equation}

\noindent\textbf{Composition loss}
The composition loss function, denoted as $\mathcal{L}_{comp}$, is computed as the absolute difference between the ground truth RGB values, $C_g^i$, and the predicted RGB values, $C_p^i$, which are obtained via alpha blending~\cite{Levin2006ACS}. Similar to the alpha loss function, a small value $\varepsilon$ is added for differentiability, and the weight $W_{c}^{i} \in \left[ 0,1 \right]$ is set following the same procedure as previously mentioned.
\begin{equation}
\mathcal{L}_{comp}= \frac{\sum_{i}\sqrt{\left ( \left ( C_p^i-C_g^i \right )\times W_c^i \right )^{2}+\varepsilon^{2}}}{\sum_{i}W_c^i}.
\label{equa:loss_comp}
\end{equation}

\noindent\textbf{MSE loss}
The MSE loss function, which stands for mean squared error, computes the average squared difference between the ground truth alpha matte, $\alpha_g$, and the predicted alpha matte, $\alpha_p$, as shown in Eq.~\eqref{equa:loss_mse}. The weight $W_{\alpha}^{i} \in \left[ 0,1 \right]$ has the same meaning as previously mentioned. Some studies~\cite{NEURIPS2018_653ac11c} have used root mean square error instead, while others~\cite{hou2019context} have computed MSE between high-level features instead of the final output. There are other works~\cite{lf,Wei2020ImprovedIM,Liu2021TowardsEF,Cheng2021DeepIM,Ding2022DeepII,Fang2022UserGuidedDH} have opted to use the alpha loss for the transition area, and the MSE loss for the foreground and background regions, or vice versa, to better leverage the benefits of each loss function.
\begin{equation}
\mathcal{L}_{mse}= \frac{\sum_{i}\left ( \left ( \alpha_p^i-\alpha_g^i \right )\times W_{\alpha}^i \right )^{2}}{\sum_{i}W_{\alpha}^i}.
\label{equa:loss_mse}
\end{equation}

\noindent\textbf{Cross Entropy loss}
For those methods that involve classification prediction for intermediate representations~\cite{cai2019disentangled,Yang2020SmartSF,Ding2022DeepII} or refinement of the auxiliary input during training~\cite{shm,lf,gfm}, the cross-entropy loss function is often used. The predicted probability for the $c$-th class of the intermediate representation is denoted as $I_p^c \in \left[0,1\right]$, while the ground truth label is represented by $I_g^c \in \left\{0,1\right\}$. The cross-entropy loss function is defined as follows:
\begin{equation}
\mathcal{L}_{ce} = -\sum_{c=1}^{C}{I_g^c}log\left(I_p^c\right).
\label{equa:loss_ce}
\end{equation}

\noindent\textbf{Laplacian loss}
The Laplacian loss function is first introduced in Context-aware~\cite{hou2019context}, which is motivated by the work \cite{niklaus2018context} and is defined as the absolute difference between two Laplacian pyramid representations. This loss function can capture local and global differences, as shown in Eq.~\eqref{equa:loss_lap}. Here, $Lap^k({\alpha_g^i})$ and $Lap^k({\alpha_p^i})$ denote the $k^{th}$ level Laplacian pyramid of the ground truth alpha map and its prediction, respectively. The contribution of each Laplacian level to the loss is scaled based on its spatial size.
\begin{equation}
\mathcal{L}_{lap} = {\sum_{i}W^i}\sum_{k=1}^{5}2^{k-1}\left \|Lap^k({\alpha_p^i})-Lap^k({\alpha_g^i}) \right \|_1.
\label{equa:loss_lap}
\end{equation}

\noindent\textbf{Gradient loss}
As mentioned in previous studies~\cite{aksoy2017designing, bmvcLutzAS18}, the above loss functions cannot adequately promote sharpness in the final output, resulting in blurred boundaries. To address this issue, Tang et al.~\cite{Tang2019LearningBasedSF} propose a gradient loss that calculates the $L^1$ distance between the gradient magnitudes of the ground truth and predicted alpha mattes, as shown in Eq.~\eqref{equa:loss_grad}. The weight $W_\alpha^i$ is set as before.
\begin{equation}
\mathcal{L}_{grad}= \frac{\sum_{i}W_\alpha^i\left\| \nabla\alpha^i_p - \nabla\alpha^i_p \right\|_1}{\sum_{i}W_{\alpha}^i}.
\label{equa:loss_grad}
\end{equation}

\noindent\textbf{Adversarial loss}
To improve the quality of the generated composite images, some methods utilize adversarial losses~\cite{goodfellow2020generative}. These methods incorporate a discriminator to differentiate between the predicted composite image and real images. The adversarial loss function is defined as shown in Eq.~\eqref{equa:loss_adv}, where $x$ is a real image, $D$ is the discriminator, and $Comp(G(x))$ is a composition function that takes the predicted alpha from the generator $G$ as input to generate a fake image.
\begin{equation}
\mathcal{L}_{adv}= log D(x)+log(1-D(Comp(G(x))).
\label{equa:loss_adv}
\end{equation}

\noindent\textbf{Multi-task loss}
Several works, such as the one proposed by Cai et al.~\cite{cai2019disentangled}, employ a multi-stream architecture to perform different sub-tasks, and thus adopt a multi-task loss to supervise the model training. Following the approach suggested by Kendall et al.~\cite{kendall2018multi}, these works define a loss function that combines different losses for each task. Specifically, the loss function, as shown in Eq.~\eqref{equa:loss_multi}, consists of two terms, namely $\mathcal{L}_{Inter}$ and $\mathcal{L}_{alpha}$, which correspond to the loss associated with the intermediate prediction and alpha prediction, respectively. Here, $\sigma_1$ and $\sigma_2$ are learnable weights to balance the two sub-tasks.
\begin{equation}
\mathcal{L}_{multi}= \frac{1}{2\sigma^2_1}\mathcal{L}_{Inter}+\frac{1}{\sigma_2}\mathcal{L}_{alpha}+log2\sigma_1\sigma_2.
\label{equa:loss_multi}
\end{equation}

\noindent\textbf{SSIM loss}
To enhance the predicted foreground structure, some studies~\cite{hatt} employ the Structural Similarity (SSIM) loss~\cite{wang2004image}, inspired by previous works~\cite{Qin_2019_CVPR,wan2018crrn}, to improve the consistency of structure in the ground truth and predicted alpha mattes. Eq.~\eqref{equa:loss_ssim} demonstrates the SSIM loss, where $\mu_p$, $\mu_g$, $\sigma_p$, and $\sigma_g$ represent the mean and standard deviation of $\alpha_p$ and $\alpha_g$, respectively.
\begin{equation}
\mathcal{L}_{ssim}= 1-\frac{(2\mu_p\mu_g+c_1)(2\sigma_{pg}+c_2)}{(\mu_p^2+\mu_g^2+c_1)(\sigma_p^2+\sigma_g^2+c2)}.
\label{equa:loss_ssim}
\end{equation}

\begin{table*}[t]
\centering
\caption{Quantitative results of traditional, auxiliary input-based, and automatic matting methods on four representative matting datasets.}
\vspace{-5pt}
\label{tab:exp_results}
\begin{threeparttable}
\resizebox{0.98\textwidth}{!}{
\setlength\tabcolsep{4pt}
\renewcommand\arraystretch{1.4}
\begin{tabular}{l|r|c|cccc|cccc|cccc|}
\hline
\rowcolor{figure_head}
   & && \multicolumn{4}{c|}{Model Complexity} & \multicolumn{4}{c|}{DIM-481~\cite{dim}} & \multicolumn{4}{c|}{alphamating.com~\cite{rhemann2009perceptually}}\\
\hline
\multirow{10}{*}{\rotatebox{90}{\textbf{Traditional}}}& Method & Pub. & Backbone & \makecell[c]{\#Params\\(M)} & \makecell[c]{Complexity\\(GMac)} & \makecell[c]{Speed (s)\\$512\times512$} & SAD & MSE & GRAD & CONN & SAD & MSE & GRAD & CONN\\
\cline{2-15}
&Bayesian~\cite{Chuang2001ABA}& CVPR'01 & \multicolumn{3}{c}{\multirow{9}{*}{\diagbox[width=5em]{}}} & 20.0852 & - & -  & -& -& 25.61 & 2.85 & 2.48 & 6.34 \\
&ClosedForm~\cite{Levin2006ACS}& TPAMI'06 &  & & & 1.8197& 168.1& 0.091& 126.9& 167.9& 17.73& 1.62 & 1.5 & 0.54 \\
&Robust~\cite{wang2007optimized}&CVPR'07 &  & & & 8.7748& -& -& -& -& 17.59& 1.48& 1.2&2.09 \\
&Learning~\cite{zheng2009learning}& ICCV'09 & & & & 3.1127 & 113.9&0.048 &91.6 &122.2 &17.3 & 1.44&1.35  &1.66\\
&Shared~\cite{gastal2010shared}&CGF'10 & & & & 98.3686& 128.9 & 0.091&126.5 &135.3 &13.55 &0.88  &0.9 &0.81 \\
&Global~\cite{he2011global}&CVPR'11 & & & & 3.5040 &133.6 &0.068 & 97.6&133.3 & 12.83 &0.81 &1.03 &1.34 \\
&Comprehensive~\cite{shahrian2013improving}&CVPR'13 & & & & 41.4687 &143.8 &0.071 &102.2 &142.7 & 12.73 & 0.75 &0.9 &1.21\\
&KNN~\cite{chen2013knn}&TPAMI'13& & & &6.6741 & 175.4& 0.103&124.1 &176.4 & 12.69 & 0.7& 1.03&0.81\\
&IFM~\cite{aksoy2017designing}&CVPR'17& & & & 15.8284& 75.4& 0.066& 63.0& -& 9.51 &0.49 &0.78 &0.71\\
\hline
\multirow{17}{*}{\rotatebox{90}{\textbf{Auxiliary}}}&DCNN~\cite{Cho2016NaturalIM}& ECCV'16 & \makecell[c]{-} & 1.51 & 8.74& 0.0017& 161.4 &0.087 & 115.1& 161.9& 10.2& 0.55 &0.79 &0.93\\
&DIM~\cite{dim}& CVPR'17 & VGG-16~\cite{Simonyan2014VeryDC} &130.55 &25.58 &0.0440 & 50.4 &0.0014 &31.0 &50.8 & 9.64 &0.57  & 0.85& 0.73\\
&AlphaGAN~\cite{bmvcLutzAS18}& BMVC'18 & ResNet-50~\cite{he2016deep}& 40.3 & 203.31&0.0364 & 52.4& 0.0030& 38.0& -& 10.2& 0.55& 0.79 & 0.93\\
&SampleNet~\cite{Tang2019LearningBasedSF}& CVPR'19 & \makecell[c]{-} & 47.83& 392.08&0.2454 &40.4 &0.0099 &-  &- & 9.47 &0.54 &0.78  &0.89\\
&AdaMatting~\cite{cai2019disentangled}& ICCV'19 & ResNet-50 & 8.49& 55.85&0.0367 & 41.7&0.0010 &16.8 &-  & 9.06 & 0.5 &0.7 &0.55\\
&IndexNet~\cite{lu2019indices}& ICCV'19 & MobileNetV2~\cite{sandler2018mobilenetv2} & 3.18&22.92 &0.0142 &45.8 &0.0013 &25.9 & 43.7& 10.31& 0.62 &0.83 &0.69\\
&Context~\cite{hou2019context}&ICCV'19& Xception 65~\cite{chollet2017xception} & 184.95 &205.47 &0.0406 &35.8 &0.0082 &17.3 &33.2 & 9.79 & 0.46& 0.64& 0.74\\
&GCA~\cite{gca}& AAAI'20 & ResNet-34~\cite{he2016deep} &25.16 &0.52 &0.0301 & 35.3&0.0091 &16.9 &32.5 & 9.59 &0.54 &0.66 &0.56\\
&FBA~\cite{Forte2020FBA}&ArXiv'20 & ResNet-50 &34.69 &7.15 &0.0425  &25.8 &0.0052 &10.6 & 20.8& -& - & - & -\\
&FDMatting~\cite{Liu2021TowardsEF}& WACV'21 & ResNet-34 & 29.0& 779.43& 0.0512& 37.6& 0.0090&18.3 &35.4 & - & - & - & -\\
&HDMatt~\cite{Yu_Xu_Huang_Zhou_Shi_2021}&AAAI'21 & ResNet-34 & 53.01& 1306.06& 0.0724& 33.5& 0.0073& 14.5& 29.9& 8.8 &0.41 &0.56 &0.69\\
&SIM~\cite{sim}&CVPR'21& ResNet-50 & 46.45& 19.88 &0.0543 &28.0 &0.0058 &10.8 &24.8 & 8.01 &0.4 &0.49 &0.51 \\
&MGMatting~\cite{mgmatting}&CVPR'21& ResNet-50 & 29.6& 4.37&0.0170 &28.9 &0.0057 &11.4 &24.9 & - & - & - & -\\
&TIMI-Net~\cite{Liu2021TripartiteIM}& ICCV'21 & ResNet-18~\cite{he2016deep} & 34.89&3.37 &0.0199 & 29.1&0.0060 & 11.5& 25.4& 7.34& 0.34& 0.45 &0.5\\
&LFPNet~\cite{Liu2021LongRangeFP}&MM'21 & ResNet-50 &117.61 &24.44 &0.1084 & 22.4 & 0.0036 & 7.6 & 17.1 & 7.67 & 0.34 & 0.4  & 0.5 \\
&FGI~\cite{Cheng2021DeepIM}&BMVC'21 & ResNet-34 & 26.04 & 2.35 & 0.0148  & 30.19& 0.0061& 13.07& 26.66&9.12 &0.47  &0.55  &0.98 \\
&MatteFormer~\cite{Park2022MatteFormerTI}& CVPR'22 & Swin~\cite{Liu2021SwinTH} & 44.75 & 29.14& 0.0545& 23.8& 0.0040& 8.7& 18.9& - & - & - & -\\

\rowcolor{figure_head}
   & && \multicolumn{4}{c|}{Model Complexity} & \multicolumn{4}{c|}{AM-2K~\cite{gfm}} & \multicolumn{4}{c|}{P3M-10K (P3M-500-NP)~\cite{p3m}}\\
\hline
\multirow{11}{*}{\rotatebox{90}{\textbf{Automatic}}}&Method &Pub. & Backbone& 
\makecell[c]{\#Params\\(M)} & \makecell[c]{Complexity\\(GMac)} & \makecell[c]{Speed (s)\\$512\times512$}  & SAD & MSE & MAD & SAD-T & SAD & MSE & MAD & SAD-T \\
\cline{2-15}
&SHM~\cite{shm} & MM'18 & ResNet-50 & 79.27&870.16&0.0993& 17.81& 0.0068& 0.0102& 10.26&20.77 & 0.0093& 0.0122& 9.14\\
&LF~\cite{lf} &CVPR'19 & DenseNet-201~\cite{huang2017densely}& 37.91& 2821.14& 0.1103& 36.12 &0.0116 & 0.0210& 19.68&32.59 &0.0131 &0.0188 &14.53 \\
&HATT~\cite{hatt} & CVPR'20 &ResNeXt-101~\cite{Xie2016AggregatedRT} &106.96 &1502.46 &0.0862 &28.01 & 0.0055& 0.0161& 13.36&30.53 &0.0072 &0.0176 &13.48 \\
&SHMC~\cite{shmc} &CVPR'20 & ResNet-34 & 78.23& 139.55& 0.0125 &61.50 &0.0270 &0.0356 & 35.23& 31.07& 0.0094&0.0179 &18.86 \\
&AIM~\cite{aim} &IJCAI'21& ResNet-34 & 55.30 & 99.56 & 0.0165 & 13.97 &0.0050 &0.0081 & 10.44& 13.01&0.0045 &0.0075 &9.24 \\
&P3M~\cite{p3m} &MM'21& ResNet-34 &39.48 & 80.1 & 0.0127 & 13.59 &0.0046 &0.0078 &10.55 & 11.23& 0.0035& 0.0065& 7.65\\
&GFM~\cite{gfm} &IJCV'22& ResNet-34 & 55.29&75.39 & 0.0135 &10.89 &0.0029 &0.0064 &9.15 &15.50 &0.0056 &0.0091 &10.16 \\
&MODNet~\cite{modnet}& AAAI'22 & MobileNetV2 & 6.49& 20.0 & 0.0131 & 19.88& 0.0052& 0.0116& 14.89&15.56 &0.0043 & 0.0090 & 9.40 \\
&P3M-ViTAE~\cite{p3mj} &IJCV'23& ViTAE~\cite{zhang2023vitaev2} & 27.46& 112.01 & 0.0399 & 9.75 & 0.0026 & 0.0057 &8.58 &7.59 &0.0019 & 0.0044& 6.60\\
\hline
\end{tabular}}
\end{threeparttable}
\vspace{-10pt}
\end{table*}

\subsection{Evaluation Metrics}
To quantitatively evaluate the quality of predicted results, previous researchers have proposed many evaluation metrics that measure prediction errors from different perspectives. We discuss some of these metrics as follows.

\noindent\textbf{SAD}
SAD is a widely used evaluation metric in the image matting task. It measures the dissimilarity between the predicted alpha matte and the ground truth alpha matte by computing the absolute difference between each pixel and summing up all the values.

\noindent\textbf{SAD-T}
Furthermore, in the context of automatic image matting, researchers have also introduced SAD-T (SAD in transition areas)~\cite{gfm}, which is a modified version of SAD that specifically measures the errors in the transition areas, which are known to contain fine details that are crucial for accurate alpha matte prediction.

\noindent\textbf{MSE} MSE measures the average squared difference between the predicted alpha matte and the ground truth.

\noindent\textbf{MAD} 
To address the limitations of SAD and MSE in not accounting for image size and being unsuitable for cross-dataset comparisons, researchers have proposed using MAD (\aka, mean absolute difference)~\cite{gfm} as an additional metric for evaluating the quality of predicted alpha mattes.

\noindent\textbf{GRAD} 
While the aforementioned metrics provide a useful foundation for comparison, they are not always indicative of the visual quality as perceived by human observers. Therefore, some researchers~\cite{rhemann2009perceptually} have proposed using gradient errors to measure the difference between the gradients of the predicted alpha matte and the ground truth matte. The gradient is obtained by convolving the alpha mattes with first-order Gaussian derivative filters with a variance.

\noindent\textbf{CONN} 
Motivated by previous works~\cite{rosenfeld1983connectivity,vincent1991watersheds}, researchers~\cite{rhemann2009perceptually} propose the CONN metric, which is similar to GRAD and aims to measure the quality of alpha mattes based on the connectivity between the foreground and background regions. Specifically, it defines the degree of connectedness based on the connectivity in binary threshold images computed from the grayscale alpha matte.

\begin{figure*}[t]
    \centering
    \includegraphics[width=\linewidth]{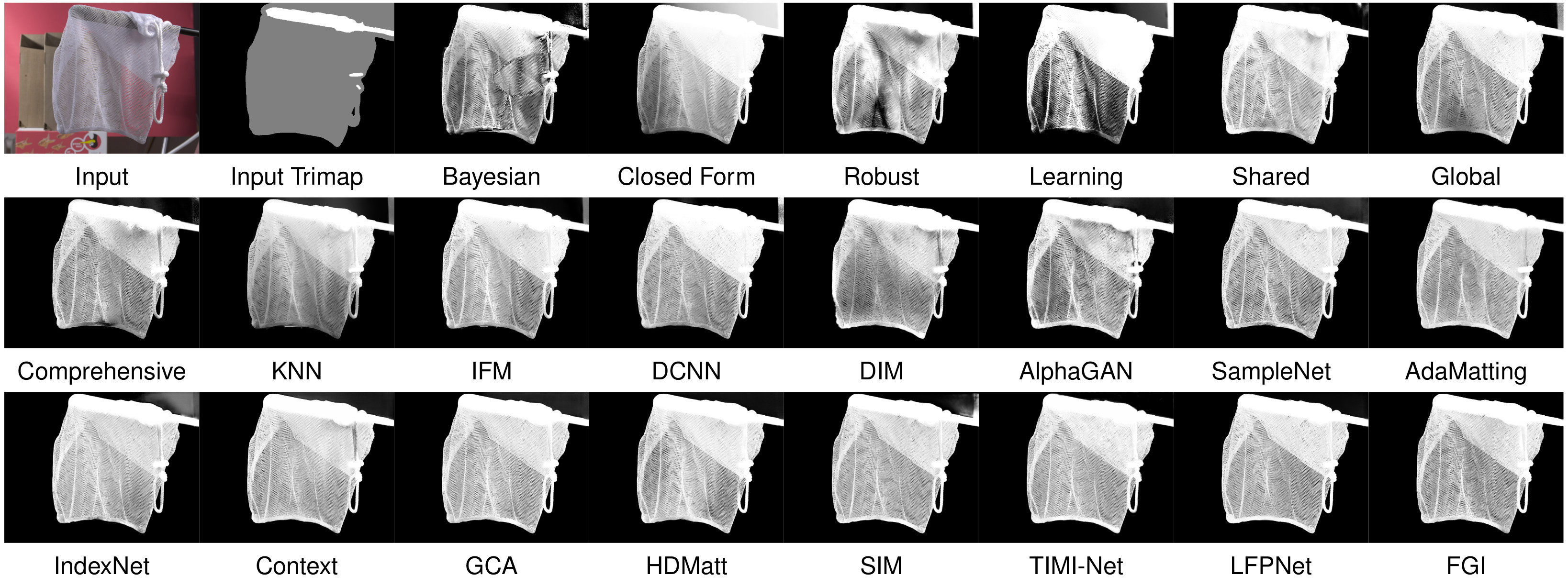}
    \caption{Some subjective results of auxiliary-based matting methods on alphamatting.com~\cite{rhemann2009perceptually}. We suggest enlarging the image for a more detailed view.}
    \label{fig:exp_auxiliary}
\end{figure*}

\begin{figure*}[t]
    \centering
    \includegraphics[width=\linewidth]{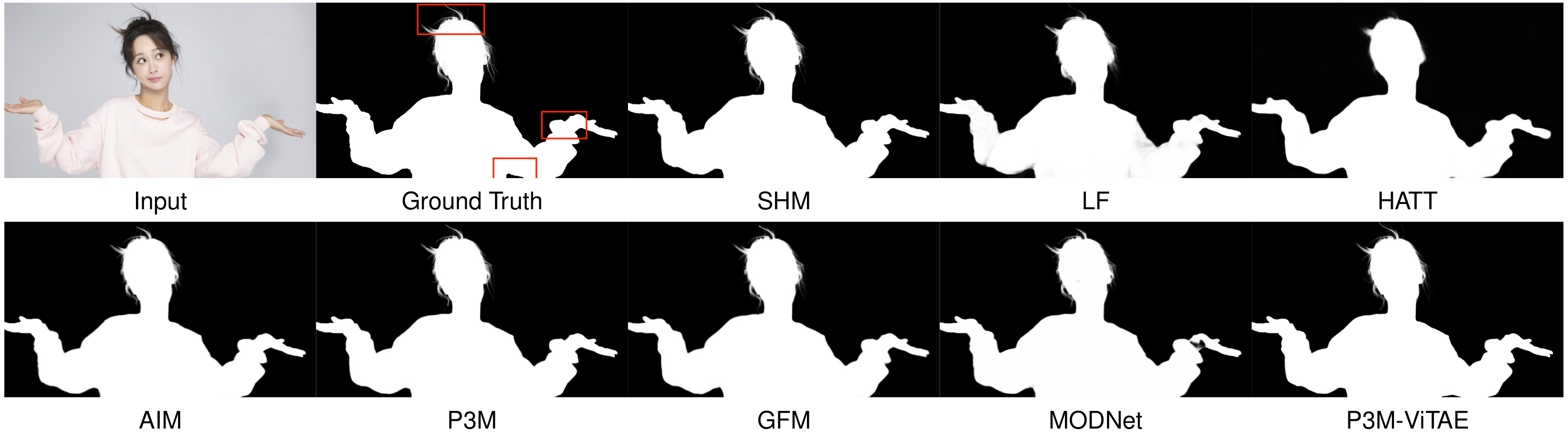}
    \caption{Some subjective results of automatic matting methods on P3M-500-NP~\cite{p3m}. We suggest enlarging the image for a more detailed view.}
    \label{fig:exp_automatic}
\end{figure*}

\subsection{Results and Analysis}
In this section, we conduct a comprehensive performance evaluation of existing image matting techniques. Specifically, we benchmark both auxiliary input-based methods and automatic methods. However, the experiment settings vary considerably due to the distinct nature of these two kinds of methods. To this end, we provide a detailed discussion of data augmentation strategies, as well as training and testing protocols. Furthermore, we present both the quantitative and qualitative evaluation results of several representative methods on four public matting datasets, as shown in Table~\ref{tab:exp_results}, Figure~\ref{fig:exp_auxiliary}, and Figure~\ref{fig:exp_automatic}.

\subsubsection{Auxiliary Input-based Image Matting}
To compare the efficacy of deep learning-based methods against traditional methods, we evaluate their performances on two widely adopted datasets: DIM-481~\cite{dim} and alphamating.com~\cite{rhemann2009perceptually} and focus on the trimap auxiliary input. Specifically, the results obtained from alphamatting.com are reported as the average scores of the prediction results based on the user-defined trimap.

\noindent\textbf{Data augmentation} 
As a common practice, the training images are typically subject to random cropping centered on pixels in the transition area~\cite{dim}, and sampled at different sizes (\eg, $320\times320$, $480\times480$, $640\times640$) before being resized to $320\times320$ to enhance the robustness of the trained models against variations in scale~\cite{dim}. Some researchers~\cite{Tang2019LearningBasedSF} have extended these strategies by using two foregrounds for composition and randomly changing the brightness, contrast, and saturation. For background image sources, MS COCO~\cite{coco} and Pascal VOC~\cite{everingham2010pascal} are commonly adopted. Recently, the use of BG-20K~\cite{gfm} has shown great advantages in terms of diversity, resolution, and quality.

As shown in the top part of Table~\ref{tab:exp_results}, the results demonstrate a significant performance gap between deep learning-based methods and traditional methods, which is consistent with the subjective results in Figure~\ref{fig:exp_auxiliary}. This can be attributed to the superior representation abilities of deep neural networks, resulting in substantial reductions in SAD from 175.4 to 22.4 on DIM-481 and 25.61 to 7.34 on alphamatting.com. The visual results in Figure~\ref{fig:exp_auxiliary} also demonstrate the superiority of deep learning-based methods over traditional ones. 

Furthermore, the results of deep learning methods reveal some interesting findings. Firstly, transformer-based models exhibit promising potential in predicting fine details, as demonstrated by their performance on DIM-481~\cite{dim}. MatteFormer~\cite{Park2022MatteFormerTI}, for instance, is a notable example that owes its success to the powerful capacity of capturing global context information. Secondly, multi-stream architectures that learn the long-range context features together with image patch details have also shown good results, \eg, TIMI-Net~\cite{tripathi2019learning} on alphamating.com and LFPNet~\cite{Liu2021LongRangeFP} on DIM-481~\cite{dim}. Thirdly, it is also important to note that using only one benchmark may result in bias since the methods may overfit a specific dataset. Hence, it is highly recommended to evaluate the performance of matting methods on various benchmarks. Finally, although the current state-of-the-art deep learning-based matting methods are already demonstrating remarkable potential, there is still room for further studies, such as improving the details of transition areas.

\subsubsection{Automatic Image Matting}
To evaluate the performance of automatic image matting methods, we choose to benchmark them on real-world public datasets, namely AM-2K~\cite{gfm} and P3M-10K~\cite{p3m}, to mitigate potential issues arising from resolution discrepancies and semantic ambiguities in composite images~\cite{gfm}. We trained the representative models according to the settings outlined in their respective papers, and in cases where code is unavailable, we re-implement it. All models were trained using the same data augmentation strategies on two NVIDIA Tesla V100 GPUs for a fair comparison.

\noindent\textbf{Data augmentation}
In contrast to trimap-based matting methods, the training images for automatic matting methods are typically cropped and sampled from the entire image instead of only focusing on the transition areas. We adopt augmentation strategies such as Gaussian noise, blur, and sharpness, as suggested by prior studies~\cite{gfm}. Additionally, brightness, color-jitter, flip, and rotation are also employed to enhance the models' generalization abilities.

In the bottom part of Table~\ref{tab:exp_results}, we report the results of representative automatic matting methods in terms of SAD, MSE, MAD, and SAD-T. These results provide valuable insights into the effectiveness of these methods and future research directions. Firstly, it is found that methods utilizing the encoder-sharing architecture tend to produce favorable results on both datasets, \eg, GFM~\cite{gfm}, AIM~\cite{aim}, and P3M~\cite{p3m}, highlighting the advantages of decomposing the matting problem into sub-tasks and addressing them jointly. Secondly, the transformer-based model such as P3M-ViTAE~\cite{p3mj}, has shown superior performance due to its remarkable ability to model long-range dependencies and extract global semantic features. Thirdly, it is noteworthy that the SAD-T metric, which measures the SAD error within the transition area, still accounts for a significant portion of the overall SAD error. This emphasizes the need for future work to focus on reducing errors in transition areas, which often contain many fine details.

\subsection{Model Complexity Analysis}
To provide a comprehensive analysis of the computational complexity and inference speed of various matting methods, we conduct evaluations on a workstation equipped with an Intel Xeon CPU (2.30GHz) and a Tesla V100 GPU. For each method, we record the average inference speed of processing a $512\times512$ image over 100 runs. As can be seen from Table~\ref{tab:exp_results}, the number of parameters ranges from 3.18M to 184.95M, with complexities varying from 0.52 GMac to 2821.14 GMac, and test speeds ranging from 0.0017 seconds to 0.2454 seconds. Notably, most auxiliary input-based matting methods and automatic matting methods can achieve real-time inference speeds. Nevertheless, there is room for further research to develop more lightweight models while maintaining good performance and fast inference speed.

\section{Applications}
\label{sec:applications}

Image matting, as a fundamental low-level image processing technique, has enabled the advancement of numerous downstream applications. These applications span a wide range, including e-commerce advertising promotion, image editing for daily entertainment, and virtual reality and augmented reality. In this section, we discuss the applications and research areas that are closely related to image matting by reviewing relevant literature.

\noindent\textbf{Visual perception}
Image matting finds applications in various visual perception tasks such as object detection~\cite{fan2021concealed, guo2012paired}, semantic segmentation~\cite{akimoto2020fast, lin2016scribblesup, ghosh2019understanding, gao2021dsp}, obstruction removal~\cite{liu2020learning}, and certain specific matting tasks such as cloud matting~\cite{zou2019generative}, medical matting~\cite{wang2021medical}, and light-field matting~\cite{cho2016automatic}. More specifically, image matting is used to refine binary shadow masks~\cite{guo2012paired} after detection, to learn to predict cloud reflectance and attenuation~\cite{zou2019generative}, and to estimate consistent mattes across light-field images based on depth and color~\cite{cho2016automatic}.

\noindent\textbf{Image editing}
Image matting has also found widespread use in downstream image editing applications such as image composition~\cite{Niu2021MakingIR, zhang2021deep}, image inpainting/outpainting~\cite{zhao2019guided, ke2023subject}, image enhancement~\cite{pandey2021total, shih2013data}, and image style transfer~\cite{luan2017deep}. Among these, image matting is often used as a common approach for image composition~\cite{zhang2021deep}, and generates alpha matte and foreground for image relighting~\cite{pandey2021total}.

\noindent\textbf{Video processing}
Image matting has also been utilized in various video-related applications, including video effects association~\cite{lu2021omnimatte}, shallow depth-of-field synthesis~\cite{lijun2018deeplens}, video style transfer~\cite{xia2021real}, multi-view stereo~\cite{yao2018mvsnet}, RGB-D matting~\cite{peng2023rgb}, and video matting~\cite{sun2021deep,zhang2021attention}. In particular, image matting contributes to generating the training data for shallow depth-of-field synthesis~\cite{lijun2018deeplens}, and extracts better portrait foreground with the aid of depth maps~\cite{peng2023rgb}.

\noindent\textbf{Multi-modality and 3D applications}
Image matting is also used in various multi-modality and 3D applications, such as remote sensing~\cite{li2020deep,pan2022remote}, 3D rendering~\cite{Lu2020LayeredNR}, and 3D matting~\cite{wang20223d}. For instance, researchers have utilized image matting for foreground extraction before 3D rendering~\cite{Lu2020LayeredNR}. Additionally, they have customized advanced matting algorithms for CT images, allowing for the calibration of the alpha matte with radiodensity~\cite{wang20223d}. In remote sensing, a deep learning-based matting framework with a clear physical significance is used to detect clouds from remote sensing images~\cite{li2020deep} and image matting has also been applied to process the intensity component obtained by averaging over each band of multispectral images~\cite{pan2022remote}.

In addition to the aforementioned applications, image matting is also being applied in various industrial and entertainment domains. For instance, it is utilized in e-commerce advertisement promotions, automatic driving scene analyses, portrait image editing, and ID photo generation from natural images. Moreover, image matting plays an important role in the metaverse and game industry, where it is used as a pre-processing technique to enable fancy effects and make the rendering process realistic and efficient.

\section{Challenges and Opportunities}
\label{sec:challenges}


In recent years, image matting has undergone rapid development with the integration of deep learning techniques. Researchers have made significant contributions by proposing new tasks, designing innovative models, improving the optimization process, and establishing large-scale datasets to advance the field. Promising results have been achieved, and some models are even capable of real-time implementation. However, there are still several challenges to overcome and opportunities to explore in this field. In the following section, we discuss the current challenges and highlight potential research directions.

\subsection{Challenges}

\noindent\textbf{Comprehensive datasets}
Establishing a challenging and large-scale dataset is crucial for developing deep learning-based image matting methods. Although researchers have constructed various training and testing datasets for image matting, these datasets are often biased towards a single category such as humans, animals, or transparent objects~\cite{dapm,gfm,cai2022transmatting}, suffer from composite artifacts due to the lack of natural images~\cite{dim,Liu2021TripartiteIM,sim}, and are not publicly available~\cite{shm,shmc,Xu2022SituationalPG}. Therefore, it is essential to construct more challenging and comprehensive datasets that contain large-scale high-resolution natural images, foregrounds from multiple categories with fine details, diverse backgrounds, and precise high-quality alpha matte annotations.

\noindent\textbf{Evaluation metrics}
Although previous researchers have designed various metrics, such as SAD and MSE, that are capable of evaluating the quality of the predicted alpha mattes in terms of their similarity with the ground truth, there may still exist discrepancies between these evaluation results and the quality perceived by humans~\cite{wang2002image}. For example, a predicted alpha matte may exhibit a small value of SAD even in the transition regions, yet fail to accurately predict fine details that are important to human observers, such as hair, earrings, or glasses frames. To address this issue, one possible solution is to adopt structural similarity-based metrics, such as SSIM, which prioritize the preservation of fine details. Nevertheless, developing new evaluation metrics that consider the characteristics of the human visual system and are suitable for image matting tasks represents a significant challenge and deserves further research efforts.

\noindent\textbf{Generalization ability}
The generalization ability of current image matting methods is significantly influenced by the training datasets. Models trained on synthetic images have been found to perform poorly on natural images~\cite{hatt,lf}, even trained with proper augmentation strategies~\cite{hou2019context,gfm,dai2022boosting}. Similarly, models trained on single-category images are unable to handle other categories of images~\cite{Chen2018TOMNetLT,mgmatting,p3m}. To overcome these limitations, it is necessary to construct more comprehensive datasets and develop advanced training methods that can improve the generalization abilities of image matting models.

\noindent\textbf{Lightweight model} 
Since image matting is typically employed in real-time industrial scenarios, advanced architecture designs are required to develop lightweight and computationally efficient models. Some effective ideas such as dimension reduction~\cite{he2016deep,sandler2018mobilenetv2}, feature reuse~\cite{huang2017densely}, token pruning~\cite{kim2021learned}, and hybrid-resolution structure~\cite{backgroundmattingv2,Yu_Xu_Huang_Zhou_Shi_2021} in related areas could be explored for efficient matting. 

\subsection{Future Prospects}

\noindent\textbf{Weakly supervised or unsupervised learning}
As the manual effort required to label high-quality matting datasets is significant, there is considerable interest in exploring training data with weak or even no annotations~\cite{zhou2018brief, jing2020self}, \eg, coarse masks in segmentation dataset~\cite{imagenet} and saliency maps in saliency detection datasets~\cite{wang2017learning, feng2022ic9600}. By using them as the pre-training data and transferring the learned knowledge obtained from these tasks to the image matting task, the matting models can obtain good weight initialization and have better generalization abilities~\cite{aim,he2022super}. In addition, off-the-shelf deep learning models designed for object detection and semantic segmentation can be leveraged to process unlabeled data and obtain object bounding boxes and masks as weak labels. These weak labels can provide valuable supervisory signals to learn semantic features for image matting models, which holds great promise for further improving their performance. Therefore, continued research in this direction is highly encouraged.

\noindent\textbf{Bridging the synthetic and real}
As previously discussed, synthetic images are of great importance in image matting due to the low cost to produce large-scale training data. Nevertheless, the use of synthetic images for training may lead to limited generalization ability due to the presence of composite artifacts~\cite{gfm}. Therefore, it is imperative to address this issue in future studies by narrowing the domain gap between synthetic and natural images. Some promising approaches to achieve this include utilizing advanced rendering or image generation techniques, developing effective data augmentation strategies, and exploring domain adaptation and generalization methods~\cite{zhang2019category}.

\noindent\textbf{Multi-modality matting}
Collaborating image matting with other modalities can extend its usage scenarios in controllable multi-modality image editing tasks. For instance, following the text instruction, RIM~\cite{rim} could generate the alpha matte of the foreground that matches the text description from a given image. Further research is warranted to investigate image matting with additional modalities, such as language~\cite{qiao2019mirrorgan,qiao2019learn}, speech, eye gaze~\cite{morimoto2005eye}, or 3D modeling based on the Neural Radiance Fields (NeRF)~\cite{Mildenhall2020NeRFRS}.

\noindent\textbf{Diffusion model}
Recently, the diffusion model~\cite{ho2020denoising} has exhibited remarkable potential in generating data, particularly images, that align well with the training data provided. The rapid advancements of large language models and cross-modality pre-training have enabled the diffusion model to comprehend human language instructions and perform image generation and editing from a blank canvas~\cite{rombach2022high,kawar2022imagic}. Now, an open question arises: \textit{What will be the impact of the diffusion model on the future of image matting, \eg, will it be a devastating setback, or will it herald a new era of possibilities?} Despite the uncertainty, we are cautiously optimistic that image matting will continue to play a crucial role in diffusion model-based image generation and editing due to its inherent flexibility, compositionality, and explainability.

\section{Conclusion}
\label{sec:conclusion}
This paper presents a comprehensive review of a substantial amount of literature on image matting. We began with an overview of the historical background and traditional techniques and examined deep learning-based methodologies, categorizing them according to the architectural designs of both auxiliary input-based and automatic image matting models. Then, we compared multiple image matting datasets and evaluated the performance of both traditional and deep learning-based image matting methods and their computing efficiency. Furthermore, we discussed the related applications, existing challenges, and future research opportunities. We believe that this study will facilitate readers in navigating this promising field efficiently and provide useful insights for future research endeavors.

\bibliographystyle{unsrt}
\bibliography{matting.bib}


\begin{IEEEbiography}[{\includegraphics[width=1in,height=1.25in,clip,keepaspectratio]{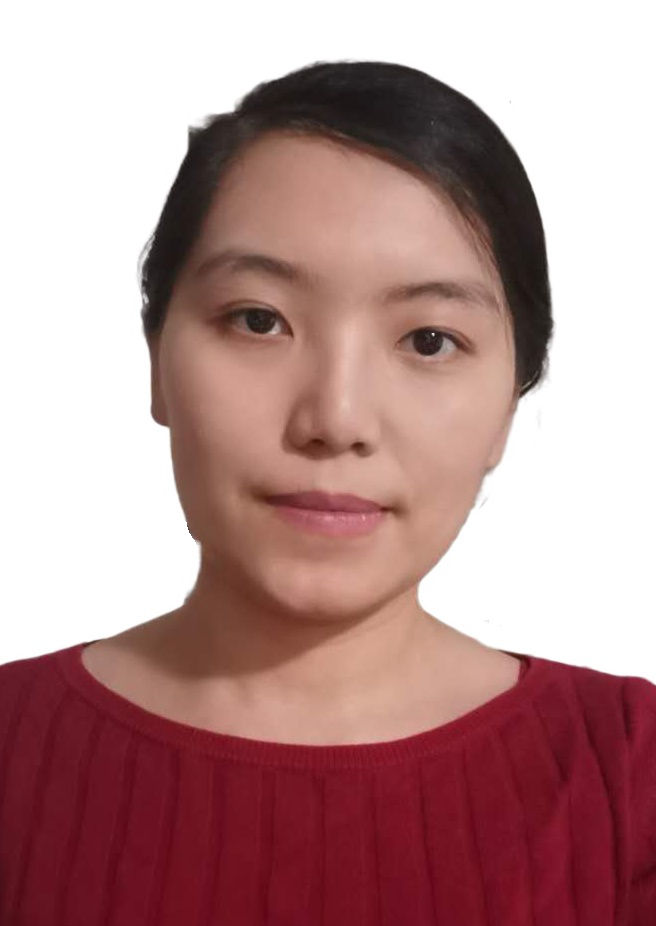}}]{Jizhizi Li} received the B.E. degree in electrical engineering from Beihang University and Master's degree in information technology from the University of Melbourne. She is currently pursuing a Ph.D. degree in the School of Computer Science from the University of Sydney. Her research interests include computer vision, image matting, and multi-modal learning. She has published several papers in top-tier conferences and journals including CVPR, IJCV, IJCAI, and Multimedia.\end{IEEEbiography} 

\begin{IEEEbiography}[{\includegraphics[width=1in,height=1.25in,clip,keepaspectratio]{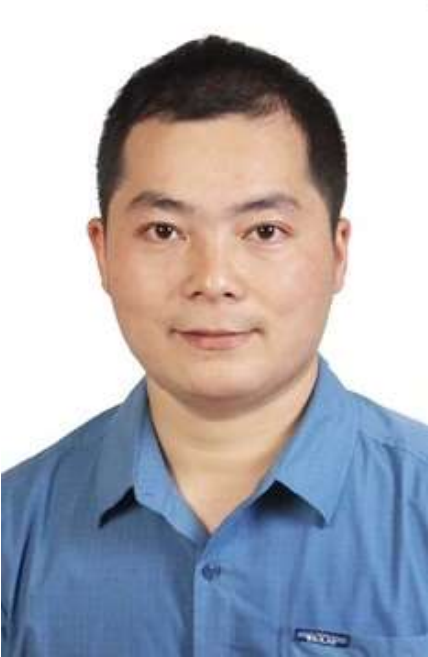}}]{Jing Zhang} (Member, IEEE) is currently a Research Fellow at the School of Computer Science, The University of Sydney. He has published more than 60 papers in prestigious conferences and journals, such as CVPR, ICCV, ECCV, NeurIPS, ICLR, International Journal of Computer Vision (IJCV), and IEEE Transactions on Pattern Analysis and Machine Intelligence (TPAMI). His research interests include computer vision and deep learning. He is a Senior Program Committee Member of AAAI and IJCAI.
\end{IEEEbiography}

\begin{IEEEbiography}
[{\includegraphics[width=1in,height=1.25in,clip,keepaspectratio]{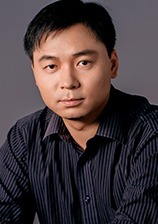}}]{Dacheng Tao} (Fellow, IEEE) is currently a professor of computer science and an ARC Laureate Fellow in the School of Computer Science and the Faculty of Engineering at The University of Sydney. He mainly applies statistics and mathematics to artificial intelligence and data science. His research is detailed in one monograph and over 200 publications in prestigious journals and proceedings at leading conferences. He received the 2015 Australian Scopus-Eureka Prize, the 2018 IEEE ICDM Research Contributions Award, and the 2021 IEEE Computer Society McCluskey Technical Achievement Award. He is a fellow of the Australian Academy of Science, AAAS, ACM, and IEEE. \end{IEEEbiography}

\end{document}